\documentclass[times,twocolumn,final,authoryear]{elsarticle}

\usepackage{ycviu}
\usepackage{framed,multirow}

\usepackage{booktabs}
\usepackage{amssymb}
\usepackage{amsmath}
\usepackage{times}
\usepackage{amsmath}
\usepackage{tikz}
\usepackage{wrapfig}
\usepackage{algorithmic,algorithm}
\usepackage{multirow}
\usepackage{graphicx}
\usepackage{subfig}
\usepackage{caption}
\usepackage{epstopdf}
\usepackage{color}
\usepackage{color}
\usepackage{cleveref}
\usepackage{latexsym}
\allowdisplaybreaks
\usepackage{xspace}
\usepackage[draft]{fixme}
\fxsetup{layout=inline,theme=color}
\definecolor{fxnote}{rgb}{0.8000,0.0000,0.0000}
\colorlet{fxnotebg}{yellow}
\makeatletter
\renewcommand*\FXLayoutInline[3]{%
	\@fxdocolon {#3}{%
		\@fxuseface {inline}%
		\colorbox{fx#1bg}{\color {fx#1}\ignorespaces #3\@fxcolon #2}}}
\makeatother

\usepackage{todonotes}

\newcommand{\LINITIAL}{\item[\algorithmiclinitial]}
\newcommand{\algorithmiclinitial}{\textbf{Initialize:}}

\makeatletter
\DeclareRobustCommand\onedot{\futurelet\@let@token\@onedot}
\def\@onedot{\ifx\@let@token.\else.\null\fi\xspace}
\def\eg{\emph{e.g}\onedot} 
\def\ie{\emph{i.e}\onedot}

\def\etal{\emph{et al}\onedot}
\makeatother

\usepackage{amssymb}
\usepackage{latexsym}

\usepackage{url}
\usepackage{xcolor}

\journal{Computer Vision and Image Understanding}

\begin{document}


\begin{frontmatter}

\title{Reliable Shot Identification for Complex Event Detection via Visual-Semantic Embedding}

\author[1]{Minnan \snm{Luo}}
\ead{minnluo@xjtu.edu.cn}
\author[2]{Xiaojun \snm{Chang}\corref{cor1}} 
\ead{cxj273@gmail.com}
\cortext[cor1]{Corresponding author}
\author[3]{Chen \snm{Gong}}
\ead{chen.gong@njust.edu.cn}
\address[1]{School of Computer Science and Technology, Xi'an Jiaotong University, Xi'an 710049, China.}
\address[2]{School of Computing Technologies, RMIT University, Melbourne, VIC 3000, Australia.}
\address[3]{Key Laboratory of Intelligent Perception and Systems for High-Dimensional Information of Ministry of Education, Nanjing University of Science and Technologyy, Nanjing 210094, China.}


\begin{abstract}
Multimedia event detection is the task of detecting a specific event of interest in an user-generated video on websites. The most fundamental challenge facing this task lies in the enormously varying quality of the video as well as the high-level semantic abstraction of event inherently. In this paper, we decompose the video into several segments and intuitively model the task of complex event detection as a multiple instance learning problem by representing each video as a ``bag'' of segments in which each segment is referred to as an instance.	Instead of treating the instances equally, we associate each instance with a reliability variable to indicate its importance and then select reliable instances for training. To measure the reliability of the varying instances precisely, we propose a visual-semantic guided loss by exploiting low-level feature from visual information together with instance-event similarity based high-level semantic feature. Motivated by curriculum learning, we introduce a negative elastic-net regularization term to start training the classifier with instances of high reliability and gradually taking the instances with relatively low reliability into consideration. An alternative optimization algorithm is developed to solve the proposed challenging non-convex non-smooth problem. Experimental results on standard datasets, \ie, TRECVID MEDTest 2013 and TRECVID MEDTest 2014, demonstrate the effectiveness and superiority of the proposed method to the baseline algorithms.  	
\end{abstract}



\end{frontmatter}


\section{Introduction}
Recent years have witnessed the unprecedented booming of multimedia data generation and distribution on the Internet thanks to the growth of platforms such as YouTube, Facebook and Twitter. 
As a natural way by which human beings interact with these multimedia data, content-based multimedia analysis is of primary importance \citep{wu2006perspectives,YuanCHLH21,ZhangCLLPH20,YanZCLYH20,ZhangLLCYH20}.
It therefore turns an interesting research efforts to content-based multimedia analysis for various applications such as multimedia information indexing and retrieval~\cite{ChengCZKK19,Zhang0LCZ18}, multimedia recommendation and multimedia event detection~\cite{ChenYZWCN20,ZhanCGCMY19}. 
As the first significant step in video analysis towards automatic categorization, recognition, search and retrieval, complex event detection that aims to automatically discover a particular event of interest in the videos, has attracting more and more research attention in the field of both computer vision and multimedia communities \citep{chang2017semantic,song2017extracting,ChangMLYH17,YanCLZZLN21,RenXCHLCW21}.

Unlike elementary visual concept detection which focus mainly on simple actions, objects, and scenes, complex event detection is a much more challenging task for the richer content and higher level semantic abstraction of the unconstrained Internet videos.
On one hand, a complex event in a video clip typically comprises of several lower level components such as multiple objects, various actions, different scenes, and the rich interactions between them; for example, \emph{``carabiners''}, \emph{``climbing gym''} and \emph{``moving hands and feet along side of rock face''} can be found in the event \emph{``rock climbing''}.
On the other hand, the quality of user-generated videos in websites varies enormously; 
In practice many video clips contain some shots that are completely irrelevant to the event of interest or even misleading \citep{vahdat2013compositional}. This makes it difficult to model these unconstrained Internet videos precisely, and consequently could potentially devastate the performance of event detection.

Technically, the key of detecting the event of interest from multimedia data lies in feature extraction and classifier training. 
Since an untrimmed video lasts for a given period of time, it is usually decomposed into several shots to capture additional local information. 
In such a way, multiple instance learning approach proposed in \citep{chen2006miles} is intuitively used for complex event detection by representing each video as a ``bag'' of segments in which each segment is referred to as an instance. 
In the framework of multiple instance learning, a video bag is labeled positive with respect to an event of interest if at least one instance in that video is positive, while the video bag is labeled negative if all the instances in it are negative. 
Note that the labels are assigned only to video bags of instances, rather than the individual instance. 
In this sense, a positive video bag often contains some instances which are irrelevant to the event, while negative bags can also contain some instances that may appear in positive bags \citep{li2015multiple}. 
As a result, there are two main issues to be considered with respect to training the classifier for complex event detection: 
\begin{itemize}
	\item How to represent an instance precisely to identify its reliability?
	\item How to alleviate the negative effect of instances with low reliability?	
\end{itemize}

To the first issue, early researches usually focus on low-lever visual features of appearance and motion in a video, such as Scale Invariant Feature Transform (SIFT) \citep{lowe2004distinctive}, Laptev’s Space-Time Interest Points (STIP) \citep{laptev2005space}, and Improved Dense Trajectory (IDT) \citep{wang2013dense,Wang2014Action,stein2017recognising}. However, these handcrafted features are  practically infeasible \citep{
chakraborty2013large}.
Leveraging on recent success in deep learning, convolution neural networks (CNN) features \citep{Karpathy2014Large} have been exploited and have yielded impressive performance. 
A complex event, however, often contains some prior knowledge, such as specific sequences or certain scenes and objects. 
Nevertheless, these visual information-based methods might fail to exploit such external information about the event of interest. 
As a result, great effort has been devoted to exploiting semantic information for multimedia event detection tasks. For example, J. SanMiguel and J. Martínez \citep{sanmiguel2012semantic} propose a framework for complex event recognition guided by hierarchical event descriptions; Concept detectors \citep{snoek2010visual} that are typically learned from different multimedia archives, have come to be leveraged to enhance the performance because the descriptions of event often contain valuable concept information \citep{habibian2014recommendations,ma2013complex,li2019zero-shot,mazloom2013searching}. 
However, these approaches heavily depend on human knowledge to design the elementary concepts space \citep{fan2017complex,cheny2019multi}; 
Moreover, the limited number of concepts that are well-defined before training manually may result in concept mis-identification during the training process.	
It is noteworthy that Venugopalan \etal \citep{Venugopalan2015Sequence} proposed a sequence-to-sequence model to describe visual content using natural language by mapping a video to a semantic description, and thereby achieve better performance.
Although research that aims to jointly exploits visual and semantic information in video is still in its early stage, these studies have demonstrated that semantic information about a video is valuable and should not be neglected.

\begin{figure}[t]
	\centering
	{\includegraphics[width=0.46\textwidth]{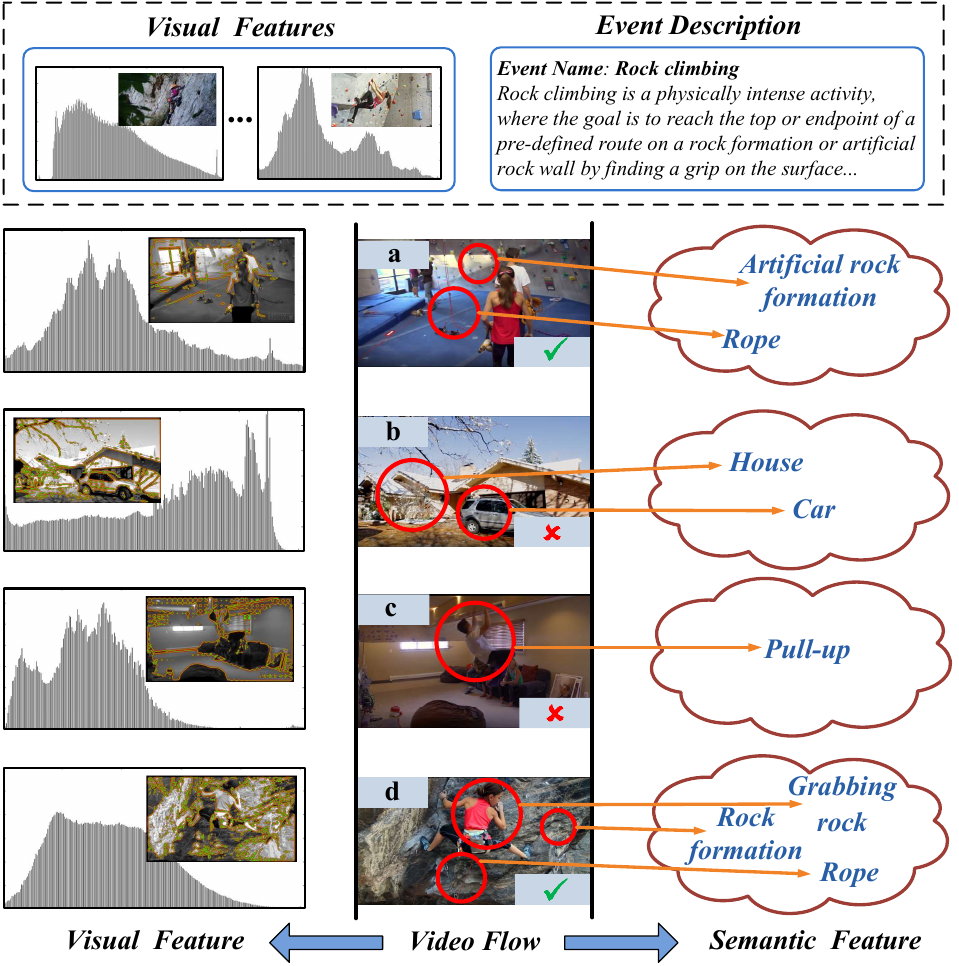}
		\caption{An example showing the semantic and visual information of different frames.\label{fig:example}}}
\end{figure}

\begin{figure*}[t]
	\centering
	{\includegraphics[width=\textwidth]{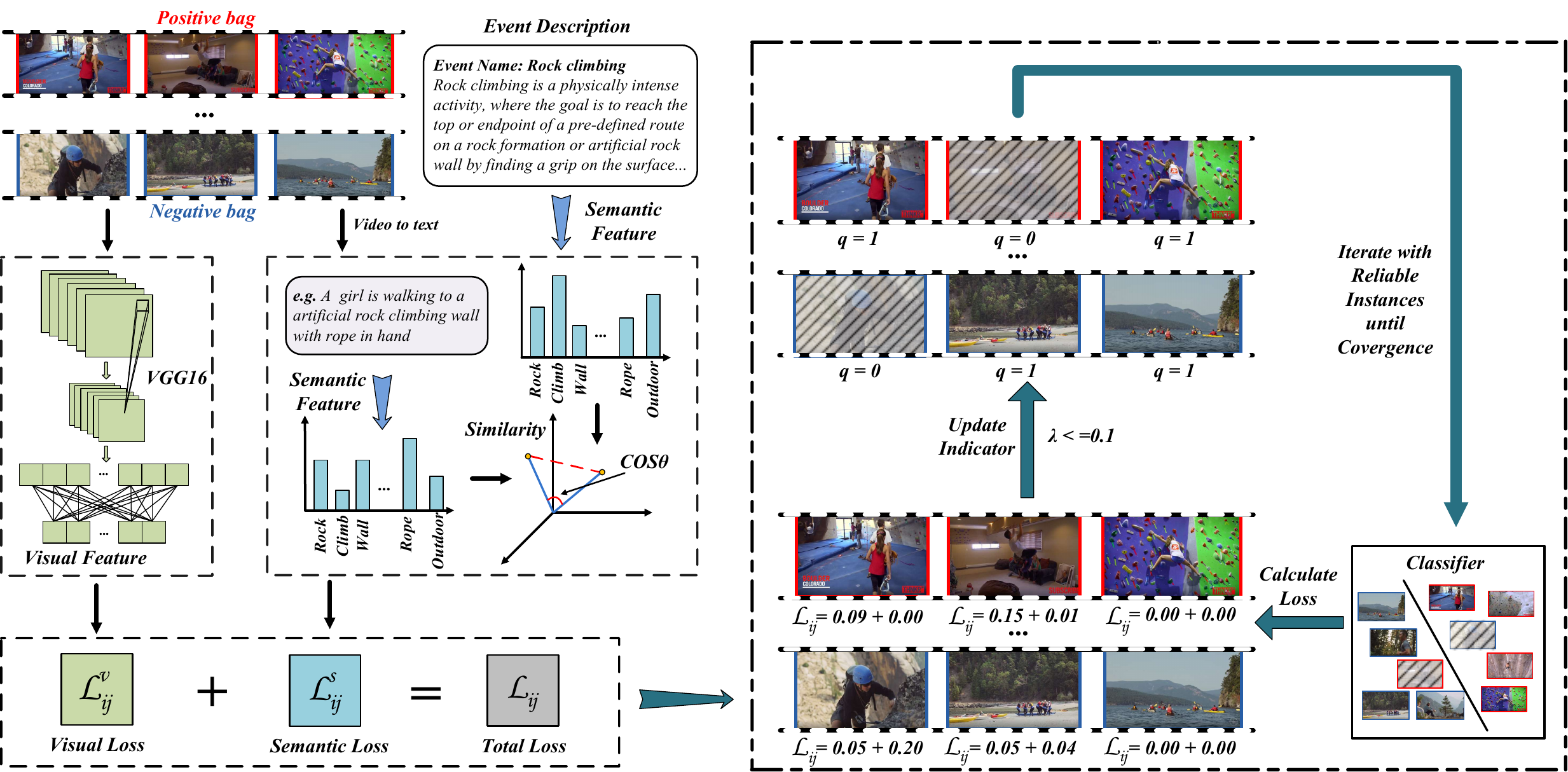}
		\caption{The framework for our training process in multimedia event detection task.\label{fig:framework}
	}}
\end{figure*}

To the second issue on the reliability of instances, Fan \etal \citep{fan2017complex} select reliable instance from positive and negative video bags by inferring a binary indicator, and train classifier on the selected reliable instances only. This strategy achieves remarkable performance on multimedia event detection task. 
\emph{However, this work neglects the semantic information involved in each instance and identifies the reliability according to visual information only, which might lead to incorrect results.} 
For example, in \Cref{fig:example}, there are some instances of a rock climbing video bag. 
It would be quite easy to identify segments \emph{b} and \emph{d} as irrelevant and reliable instances respectively by looking at visual or semantic information alone. The instance \emph{a} may be classified as irrelevant since there is no visual feature pertaining for the event \emph{rock climbing}. 
However, the concepts \emph{``artificial rock formation''} and \emph{``rope''} found as tagged in the red circle in instance \emph{a} are relevant to the event \emph{rock climbing} if semantic information is taken into consideration. 
In this sense, instance \emph{a} should be determined as a reliable instance and have a positive effect in the training stage. 
On the other hand, although instance \emph{c} contains an action feature that resembles climbing (``pulling up''), it does not contain any semantic concept of the event \emph{rock climbing}. Indeed, this instance may appear in many video bags of other events, such as \emph{``physical training''} or \emph{``indoor sport''} video bags. As a result, this instance should not be regarded as reliable in \emph{rock climbing} video bag.

\Cref{fig:framework} illustrates an overview of our proposed training process of multimedia event detection.
In our work, we take both the high-level semantic information (similarity between the concept of event description and the semantic feature of the instance own) and low-level visual information (CNN feature of the instance) into consideration to learn a reliability variable for each instance in order to indicate its importance (see the left part of \Cref{fig:framework}).
Since the instances with low reliability are difficult to use for training a robust classifier, we are motivated by \emph{curriculum learning} \citep{bengio2009curriculum} and start training the classifier on high-reliability instances first, and then gradually take the instances with relatively low reliability into consideration (see the right part of \Cref{fig:framework}). 
We formulate our proposed approach as an optimization problem, which turns out to be highly non-convex, and hence we propose an alternating algorithm to search for the optimal value of the reliability variables and classifier parameters simultaneously. 
The highlights of our work are summarized as follows:
\begin{itemize}
	\item Taking the visual low-level feature and high-level semantic information simultaneously, we propose a visual-semantic guided loss to measure the reliability of instance in the framework of multi-instance learning for event detection. 	
	\item To alleviate the negative influence of irrelevant and ambiguous segments in the training process, we 
	begin the training with high-reliability instances and gradually added in instances with relatively low reliability over time.	
	\item We conduct extensive experiments on two standard datasets, \ie, TRECVID MEDTest 2013 and TRECVID MEDTest 2014. The promising results demonstrate the effectiveness and superiority of the proposed method to the state-of-the-art methods.
\end{itemize}

The remainder of this paper is organized as follows. 
In Section \uppercase\expandafter{\romannumeral2}, we give a brief review of some related works on multimedia event detection. Our visual-semantic guided reliable shot identification for complex event detection is proposed in Section \uppercase\expandafter{\romannumeral3}.
An efficient algorithm is presented in Section \uppercase\expandafter{\romannumeral4} for finding the solution.
In Section \uppercase\expandafter{\romannumeral5}, extensive experiments over benchmark datasets are conducted to verify the effectiveness and superiority of the proposed algorithm. Finally, conclusions are given in Section \uppercase\expandafter{\romannumeral6}.

\noindent\textbf{Notations and Terms:}
Throughout this paper, we follow the standard notation and use normal lowercase characters for scalars (\eg, $z\in\mathbb{R}$), bold lowercase characters
for vectors (\eg, $\mathbf{z}=[z_1,z_2,\cdots,z_d]^\top\in\mathbb{R}^d$), normal uppercase characters for matrices (\eg, $Z=[\mathbf{z}_1,\mathbf{z}_2,\cdots,\mathbf{z}_p]\in\mathbb{R}^{d\times p}$), and calligraphic alphabets for sets (\eg, $\mathcal{Z}$).
The transpose of matrix $Z$ is denoted by $Z^\top$. 
The $\ell_{2}$-norm and $\ell_{1}$-norm of vector $\mathbf{z}\in\mathbb{R}^d$ are defined as $\|\mathbf{z}\|_{2}=\sqrt{\sum_{i=1}^{d}z_i^2}$ and $\|\mathbf{z}\|_{1}=\sum_{i=1}^{d}|z_i|$, respectively.

\section{Related Work}
In this section, we briefly review the existing related works which are relevant to multimedia complex event detection, multi-instance learning, and self-paced learning.

\subsection{Multimedia Event Detection}

It is crucial and difficult to represent the videos precisely for multimedia event detection since the long videos like those from TRECVID MEDTest-13 and TRECVID MEDTest-14 usually comprise of several lower level components such as multiple objects, various actions, different scenes, etc. 
Early research basically extracts and aggregates various complementary low-level feature descriptors from the whole video to create a unique vector representation~\cite{ZhangHJYC17,MaCYSH17,ChangY17,ChangYYX16}. 
To name a few, Shen \etal \citep{shen2008modality} leverage multimodal information and apply subspace selection technique to generate video descriptor; Sun \etal \citep{sun2013large} use Fisher Vector coding \citep{sanchez2013image} as a robust feature pooling technique to combine four types of descriptors, \ie, motion boundary histogram (MBH) \citep{wang2013dense}, histograms of gradients (HoG), optical flow (HoF) and the shape of the trajectories;
Oneata \etal \citep{oneata2013action} combine three feature descriptors including dense MBH \citep{wang2013dense}, SIFT and mel-frequency cepstral coefficients (MFCC) audio features \citep{rabiner2007introduction} with Fisher vector encoding for characterizing complex event detection task, and come to a conclusion that SIFT and MFCC features provide complementary cues for complex events. 
Xian \etal \citep{xian2016evaluation} use a uniform experimental setup to evaluate seven different types of low-level spatio-temporal features in the context of surveillance event detection. Despite of their good performance, the low-level features fail to capture the inherent semantic information in an event. 

Concept detectors that utilize several external image/video archives and learn a high-level semantic representation for the videos with complex contents \citep{snoek2010visual}, are exploited to advance event detection for its consistence with human’s understanding and reasoning.
For example, Natarajan \etal \citep{natarajan2012multimodal} combine a large set of features from different modalities using multiple kernel learning and late score level fusion methods, where the features consists of several low-level features as well as high-level features obtained from object detector responses, automatic speech recognition, and video text recognition.
Jiang \etal \citep{jiang2012leveraging} train a classifier from low-level features, encode high-level feature of concepts into graphs, and diffuse the scores on the established graph to obtain the final prediction of event.
To mitigate the unavoidable noise in concept high-level features, Yan \etal \citep{yan2015event} select the high-level semantic meaningful concepts based on both events-kit text descriptions and concept detectors, and learn a concept-driven event oriented dictionary representation for complex event detection; 
Chang \etal \citep{chang2016bi} weight the semantic representations attained from different multimedia archives and propose a semantic representation analyzing framework on both source-level and the overall concept-level. 
Due to the growth of deep convolution neural networks (CNN) \citep{krizhevsky2012imagenet}, CNN descriptors have been exploited for multimedia event detection and achieved impressive performance improvements \citep{xu2015discriminative,zha2015exploiting}.
However, these traditional approaches typically extract and aggregate local descriptors from video frames or shots to create a unique vector representation for the entire video. 
This strategy might fail to make full use of the important structural or temporal information contained in the videos \citep{zhao2018complex}, such that the key evidences are diluted for event detection, especially when the event of interest only occurs within a short period of time in an untrimmed long video. 

To tackle the issues mentioned above, several research are devoted to the efforts to exploit the evidences of event interest for better performance of event detection. 
To name a few, Tang \etal \citep{tang2012learning} divided video into several segments
and discovered the discriminative and interesting segments by leaning latent variables over the frames based on the variable-duration hidden Markov model;
Lai \etal \citep{lai2014video} represented each video as multiple ``instances'' with different temporal intervals, and inferred the instance labels and the instance-level classifier simultaneously. 
Fan \etal \citep{fan2017complex} also followed the multi-instance learning framework
and estimated a linear SVM classifier together with the selection procedure of reliable training instances.
Note that these approaches focus on the visual information contained in each instance (segment) and ignore the semantic information.
As a result, Chang \etal \citep{chang2017semantic} prioritized the segments according to their semantic saliency scores which assess the relevance of each shot with the event of interest, and then developed a nearly-isotonic SVM classifiers to exploit the constructed semantic ordering information.
Phan \etal \citep{phan2015multimedia} measured the importance of each segment by matching its detected concepts against the evidential description of the event interest, and jointly optimized with instance visual feature in a variant of multiple instance learning framework.

\subsection{Multi-instance Learning}
Multi-instance learning was first proposed in \citep{dietterich1997solving} and has been applied in several domains successfully, such as image categorization \citep{chen2004image}, object detection \citep{zhang2006multiple}, drug activation prediction \citep{wang2019bag}, and retrieval \citep{zhang2002content}.
In the framework of multi-instance learning, an example is regarded as an instance, while a bag labeled as positive or negative is composed of several instances.
Specifically, a positive bag is defined as containing at least one positive instance, while a negative bag contains no positive instances. 
The classifier is finally designed to classify bags, rather than individual instances. Note that the label of a bag can be assigned easily once all instances have been labeled. It is noteworthy that Tibo \etal \citep{tibo2020learning} introduced multi-multi instance learning for particular way of interpreting predictions, where examples are organized as nested bags of instances.
Various works have been published about bag representation by merging instances \citep{gartner2002multi}, the instance distributions of bags \citep{bunescu2007multiple}, and the relation between multi-instance learning and semi-supervised learning \citep{zhou2009multi}; however, these methods are based on the constraint that a positive bag can be determined by the existence of at least one positive instance. This assumption leads to a lack of analysis of other positive instances and is too strict for negative bag \citep{li2011text}. Since negative bags may contain several positive instances in many tasks, some works that focus on relaxing the above constraint have been developed. 
For example,  Li \etal \citep{li2011text} established a general constraint that a positive bag should contain at least a certain percentage positive instance. 
Moreover, considering that the bag can be represented by key instances, a clustering algorithm has been applied to detect these key instances \citep{liu2012key}. 
Li and Vasconcelos \citep{li2015multiple} showed that using the most positive-liked $k$ instances can result in better performance. However, the instance labels are updated under weak supervision, which may lead to unreliable solutions \citep{zhang2015self}. 
For multimedia event detection task, a video is regarded as a bag and the segments of the video are treated as instances, after which the classifier is trained on the instances. Intuitively, the inclusion of irrelevant and ambiguous segments may have a negative influence on the training classifier. However, there has been limited research on training classifiers using only those instances with strong correlations to an event for the complex event detection tasks \citep{fan2017complex,LiNCYZS18,LiNCNZY18,LuoNCYHZ18}.

\subsection{Self-pace Learning}
Inspired by the learning mode of human beings, curriculum learning \citep{bengio2009curriculum} and self-pace learning \citep{kumar2010self-paced,LuoCLNHZ17} were proposed to learn from easy samples to hard samples in training process to alleviate the negative effect of noisy samples. 
Different from curriculum learning based on certain easiness measurements, self-pace learning can automatically and dynamically choose the training order of all samples during training process \citep{Jiang2014Self}. Original self-paced learning is focus on the easiness of samples, Jiang \etal. \citep{Jiang2014Self} proposed an approach called self-paced learning with diversity which formalizes the preference for both easiness and diversity of samples via a non-convex negative $\ell_{2,1}$-norm. Self-pace learning has been widely applied to various fields, such as object tracking \citep{supancic2013self-paced}, image classification \citep{tang2012self-paced}, and multimedia event detection \citep{jiang2014easy}. Self-paced learning mode can also be integrated to many existing frameworks to enhance the performance. Zhang \etal \citep{zhang2017co-saliency} combine the multi-instance learning problem with self-pace learning to improve the performance in co-saliency detection.  Li \etal \citep{li2016multi-objective} proposed a multi-objective method to enhance the convergence of the self-pace learning algorithms. Zhao \etal \citep{zhao2015self-paced} introduced a soft self-paced regularizer to matrix factorization to impose adaptive weights to samples. 
Zhou \etal \citep{zhou2018deep} applied self-paced learning framework into deep learning to learn the stable and discriminative features.  
Huang \etal \citep{huang2019similarity} developed a similarity-aware network representation learning based on self-paced learning by accounting for both the explicit relations and implicit ones. Dizaji \etal \citep{ghasedi2019balanced} exploited a balanced self-paced learning algorithm for deep generative adversarial clustering network.

\section{The Proposed Methodology}
\label{sec:proposed}

In this section, we first introduce a visual-semantic guided loss measure at the instance level, and then propose a multi-instance learning based reliable shot identification model for multimedia event detection tasks. 

In this paper, multimedia event detection task regarding event $e$ is formulated as a binary classification problem in the framework of multi-instance learning, where the event $e$ usually comes with a short textual description in most video event datasets such as TRECVID MED13 and TRECVID MED14. 
Formally, we use the skip-gram neural network model \citep{mikolov2013distributed} in natural language processing to convert the textual description of event $e$ into a vector representation $\mathbf{e}\in\mathbb{R}^d$.
Suppose there are $n$ video bags for the detection of event $e$, denoted by $\mathcal{B}=\{(\mathcal{B}_i,y_i):y_i \in \{-1,1\};i=1,2,\cdots,n\}$, where $\mathcal{B}_i$ refers to the $i$-th untrimmed long video which are partitioned as a set of video shots (instances), \ie, $\mathcal{B}_i=\{I_{ij}:j=1,2,\cdots,m_i\}$ for $i=1,2,\cdots,n$. 
If the $i$-th video bag $\mathcal{B}_i$ belongs to the event $e$, $y_i=1$; otherwise, $y_i=-1$.
Without loss of generality, we initialize the instance label $y_{ij}$ as the corresponding bag label, \ie, $y_{ij} = y_i$ for $i=1,2,\cdots,n$ and $j=1,2,\cdots,m_i$. 
We extract the CNN features from a set of uniformly sampled frames within each video shot $I_{ij}$ to represent this video shot as $\mathbf{x}_{ij}\in\mathbb{R}^p$. 
To exploit the semantic information contained in the video shots, we generate a text description $\mathbf{t}_{ij}\in\mathbb{R}^d$ for each video shot $I_{ij}$ using an end-to-end sequence-to-sequence model proposed in \citep{Venugopalan2015Sequence}. 

\subsection{Visual-semantic Guided Loss Measure}
To make better use of the semantic and visual information of a video simultaneously, we intuitively define the visual-semantic guided loss measure on each instance $I_{ij}$ as a convex combination of losses with regard to semantic and visual information respectively, \ie,
\begin{align}
\label{Lij}
\mathcal{L}_{ij}=\alpha\mathcal{L}^v_{ij}+(1-\alpha)\mathcal{L}^s_{ij}
\end{align}
for $i=1,2,\cdots,n$ and $j=1,2,\cdots,m_i$, where $\alpha\in[0,1]$ controls the influence of losses on visual and semantic information. 
Specifically, the visual information loss on the video shot $I_{ij}$ is calculated based on hinge loss function by
\begin{align} 
\label{matirx6}
\mathcal{L}^v_{ij}(\mathbf{w},b;\mathbf{x}_{ij},y_{ij})=\max(0, 1-y_{ij}(\mathbf{w}^T\mathbf{x}_{ij}+b))
\end{align}
where $\mathbf{w}$ and $b$ are parameters to learn.  

The instances exhibiting higher correlation at the semantic level are more important to the event \citep{phan2015multimedia} than the ones with lower correlation. 
As a result, we measure the similarity between the semantic feature of each video shot $I_{ij}$ and the event of interest $e$ by
\begin{align*}
s_{ij}^{e}=\cos(\mathbf{t}_{ij},\mathbf{e})
\end{align*}
for $i=1,2,\cdots,n$ and $j=1,2,\cdots,m_i$, namely the instance-event similarity.
Note that this similarity measurement is different from the one defined in \citep{phan2015multimedia} which is formulated based on limited number of concepts.
With the instance-event similarity, the label of instance $I_{ij}$ is predicted by function $h_r:[0,1]\rightarrow \{-1,1\}$ with respect to a related level threshold $r\in\mathbb{R}$ for all video shots, \ie,
\begin{align}
\label{matirx3}
h_r(s^e_{ij})=\left\{ \begin{array}{ll}
1,  & \textrm{if $Rank(s^e_{ij})\leq r$}\\
-1,  & \textrm{otherwise}
\end{array} \right.
\end{align}
where the function $Rank(s^e_{ij})$ is utilized to quantifies the similarity $s^e_{ij}$ into a related level. 
For each instance $I_{ij}$, the value of $Rank(s^e_{ij})$ being less than $r$ means that the instance can be predicated as having a positive semantic similarity level with high confidence. It is evident that this confidence increases as the value of threshold $r$ decreases.
We define the semantic information loss $L^s_{ij}$ by penalizing the noisy instances, \ie,
\begin{align}
\label{matirx4}
\mathcal{L}^s_{ij}\left(h_r(s^e_{ij}),y_{ij}\right)
&=\left\{\begin{array}{ll}y_{ij}(1 - 2s_{ij}^{e}) + 1,  & h_r(s^e_{ij})\neq y_{ij}\\0, &\textrm{otherwise}\end{array}\right.
\end{align}
Note that the loss of correctly labeled instance is set as $0$ according to \Cref{matirx4}. For the wrongly labeled instances, moreover, we calculate loss on the following cases:
\begin{itemize}
	\item When the wrongly labeled instance $I_{ij}$ is in a positive bag, its semantic loss $\mathcal{L}^s_{ij}$ turns to $2 - 2s_{ij}^{e}$, which means we can apply a higher penalty to those instances with lower similarity to the event of interest.
	\item When the wrongly labeled instance $I_{ij}$ is in a negative bag, its semantic loss $\mathcal{L}^s_{ij}=2s_{ij}^{e}$, which suggests we can apply a higher penalty to those instances with higher similarity to the event of interest.
\end{itemize}

\subsection{Reliable Shots Identification in Multi-instance Learning Framework}
To characterize the importance of instances with different reliability, we introduce a variable $q_{ij}\in \{0, 1\}$ for each instance $I_{ij}$, and collect the reliability variables of video bag $\mathcal{B}_i$ into $\mathbf{q}_i = [q_{i1}, q_{i2}, \cdots, q_{im_i}]^\top\in\{0, 1\}^{m_i}$ for $i=1,2,\cdots,n$.
Intuitively, the instance $I_{ij}$ accrues more reliability when the value of $q_{ij}$ gets closer to $1$.
To identify the reliable instances for multimedia event detection task, we assign nonzero weights to reliable instances on the one hand and disperse these instances across more bags on the other hand. 
Following the strategy used in \citep{fan2017complex}, we learn the latent reliability variables $\{\mathbf{q}_i\}_{i=1}^n$ and the classifier parameters $\mathbf{w}, b$ jointly by minimizing the weighted training loss together with the elastic-net regularization term, \ie
\begin{align}
\label{matirx8}
&\min_{\mathbf{w}, b, \{\mathbf{q}_i\}_{i=1}^n} \sum_{i=1}^{n}(\mathbf{q}_{i}\odot\mathcal{L}_i(\mathbf{w}, b;\alpha,r) + \Omega(\mathbf{q}_i, \lambda, \gamma\mathbf))\\
&\ \ \ \ s.t.\ \ \ \ \ \sum_{j=1}^{m_i}q_{ij}\geq p_im_i\ (i=1,2,\cdots,n)\notag\\
&\ \ \ \ \ \ \ \ \ \ \ \ \ q_{ij}\in \{0, 1\}\ (i=1,2,\cdots,n;j=1,2,\cdots,m_i)\notag
\end{align}
where ``$\odot$'' denotes the element-wise product; The first constraint refers that the proportion of reliable instances in each video bag is not less than $p_i\in\mathbb{R}$ $(i=1,2,\cdots,n)$. $\mathcal{L}_i(\mathbf{w}, b;\alpha,r)=\left[\mathcal{L}_{i1},\mathcal{L}_{i2},\cdots,\mathcal{L}_{im_i}\right]^\top\in\mathbb{R}^{m_i}$ represents the visual-semantic guided loss measurement on the $i$-th video bag, where its $j$-th component is calculated by
\begin{align}
\label{lossij}
\mathcal{L}_{ij}
=\alpha\mathcal{L}^v_{ij}(\mathbf{w},b;\mathbf{x}_{ij},y_{ij})+(1-\alpha)\mathcal{L}^s_{ij}\left(h_r(s^e_{ij}),y_{ij}\right)
\end{align}
for $i=1,2,\cdots,n$ and $j=1,2,\cdots,m_i$. 
The regularization term $\Omega(\mathbf{q}_i, \lambda, \gamma)$ is defined as an negative elastic-net regularization term combining the $l_1$-norm and $l_2$-norm, \ie,
\begin{align}
\Omega(\mathbf{q}_i, \lambda, \gamma)= - \lambda\|\mathbf{q}_i\|_1 - \gamma\|\mathbf{q}_i\|_2
\end{align}
for $i=1,2\cdots,n$, where the parameters $\lambda$ and $\gamma$ are imposed on the reliability term and diversity term, respectively. Specifically, on one hand, the reliability term $\lambda\|\mathbf{q}_i\|_1$ tends to assign nonzero weights to the instances with high reliability over the instances with relatively low reliability. However, when $\lambda \neq 0$ and $ \gamma=0$, the selected instances may come from only specific bags, which may lead to overfitting. 
On the other hand, the diversity term $\gamma\|\mathbf{q}_i\|_2$ tends to assign nonzero weights  to diverse instances residing in more bags. When $\lambda = 0$ and $ \gamma \neq 0$, the algorithm selects only diverse instances so that some noisy instances may be selected, which may make the model to become biased.


\section{Optimization Strategy}
The objective function of optimization problem \Cref{matirx8} is non-convex and non-smooth, thus it is difficult to find the global minimum. In this section, we exploit an efficient alternative optimization algorithm to address this challenging problem.

To make the optimization problem \Cref{matirx8} tractable, we set the related level $r$ within the range of $\{1,2,\cdots,10\}$ and choose the best performance. Considering the computational complexity, we initialize the instance label $y_{ij}$ to be the same as the label of its corresponding video bag $\mathcal{B}_i$ for $i=1,2,\cdots,n$ and $j=1,2,\cdots,m_i$.
The reliability variable $q_{ij}$ is initialized as $1$ for every instance $I_{ij}\ (i=1,2,\cdots,n$ and $j=1,2,\cdots,m_i)$.

\paragraph{\textbf{Update $(\mathbf{w},b$)}} 

This step aims to update the multi-task network parameters with the fixed reliability  variable $Q$ over the video set $\mathcal{B}$. 
Note that the instance-event similarity $s_{ij}^e$ is fixed for each instance $I_{ij}$, and therefore the semantic loss $\mathcal{L}^s_{ij}$ becomes a constant after $y_{ij}$ is fixed according to \Cref{matirx4}. Moreover, the elastic-net regularization term $\Omega(\mathbf{q}_i, \lambda, \gamma)$ also becomes a constant. 
As a result, the optimal parameter $(\mathbf{w},b)$ is obtained by solving the following optimization problem
\begin{align}
\label{matirx10}
&\min_{\mathbf{w}, b}\sum_{i=1}^{n}\sum_{j=1}^{m_i}q_{ij}\max(0, 1-y_{ij}(\mathbf{w}^T\mathbf{x}_{ij}+b))
\end{align}
It is evident that this problem can be easily solved as a classic weighted SVM problem through using some pre-computed kernel technique. 

\paragraph{\textbf{Update $\{\mathbf{q}_i\}_{i=1}^n$}} 
With fixed variables $\mathbf{w}$ and $b$, the optimal indicator matrices $\{\mathbf{q}_i\}_{i=1}^n$ corresponding to $n$ video bags $\mathcal{B}_i\ (i=1,2,\cdots,n)$ can be learned individually. Specifically, for the $i$-th video bag $\mathcal{B}_i$, the visual-semantic guided loss of its $j$-th instance turns to a constant $\mathcal{L}_{ij}$. 
As a result, its reliable vector $\mathbf{q}_{i}$ over $m_i$ instances in the $i$-th video bag $\mathcal{B}_i$ is learned by solving the following optimization problem
\begin{align}
\label{matirx11}
&\min_{\mathbf{q}_{i}\in\{0,1\}^{m_i}}\sum_{j=1}^{m_i}q_{ij}\mathcal{L}_{ij}- \lambda\|\mathbf{q}_i\|_1 - \gamma\|\mathbf{q}_i\|_2\\
&\ \ \ \ s.t.\ \ \ \ \ \sum_{j=1}^{m_i}q_{ij}\geq p_im_i\notag
\end{align}	
for $i=1,2,\cdots,n$. We follow the algorithm proposed in \citep{Jiang2014Self} to achieve the global optimum of this non-convex problem. 
Specifically, we ascend the instance losses in the $i$-th video bag with fixed variable $\mathbf{w},b$ and let $g(t)$ be the corresponding index mapping such that 
\begin{align}
\mathcal{L}_{it}^{'}:=\mathcal{L}_{ig(t)}
\end{align}
for $t=1,2,\cdots,m_i$. 
In this sense, the $g(t)$-th instance will be selected as a reliable one for training if the following inequality
\begin{align}
\label{matirx12}
t\leq p_im_i  \ or \ \mathcal{L}_{it}^{'} < \lambda+ \frac{\gamma}{\sqrt{t}+\sqrt{t-1}}
\end{align}
holds, \ie, the reliability variable $q_{ig(t)}$ will be reset as $1$ and vice versa.
It is noteworthy that, on one hand, since the rank $t$ has a value within the range 1 to $m_i$ for the $i$-th video bag, the constraint $t\leq p_im_i$ guarantees that at least $p_im_i$ instances are selected as reliable ones. 
On the other hand, the threshold $\lambda+ \frac{\gamma}{\sqrt{t}+\sqrt{t-1}}$ decreases when the rank $t$ increases for each video bag, which can make selected instances comes from more different bags to keep diversity \citep{Jiang2014Self}. 
\begin{algorithm}[t]
	\caption{Optimization Procedure}
	\begin{algorithmic}[1]
		\REQUIRE Instance feature$\{\mathbf{x}_{ij}\}$;
		Instance-similarity $s^e_{ij}$;
		Video-level label$\{Y_{i}\}_{i=1}^n$;
		Reliability ratio $\{p_{i}\}_{i=1}^n$;
		Parameters $\alpha$, $\lambda$, $\gamma$, $r$. 
		\ENSURE $\mathbf{w}$, $b$ and $\mathbf{q}_i\ (i=1,2,\cdots,n)$.
		\LINITIAL $q_{ij}\leftarrow 1$,  $y_{ij}\leftarrow y_i$ $(i=1,2\cdots,n;j=1,2,\cdots,m_i)$. 
		\FOR{$r=1$ to $10$}
		\WHILE{not converge}
		\STATE Optimize ($\mathbf{w}$, $b$) by SVM with fixed $\mathbf{q}_i\ (i=1,2,\cdots,n)$.
		\FOR{$i=1$ to $N$}
		\STATE Calculate loss $\mathcal{L}_{i}(\mathbf{w}, b,\alpha,r)\in\mathbb{R}^m_i$ according to \Cref{lossij} and sort its components in ascending order $\rightarrow \mathcal{L}_{i}^{'}$.
		\STATE Set $g(t)$ as the index mapping such that $\mathcal{L}_{it}^{'} =\mathcal{L}_{ig(t)}$ $(t=1,2,\cdots,m_i)$.
		\FOR{$t=1$ to $m_i$}
		\IF{$t\leq p_im_i  \ \text{or} \ \mathcal{L}_{it}^{'} < \lambda+ \frac{\gamma}{\sqrt{t}+\sqrt{t-1}}$}
		\STATE $q_{ig(t)}\leftarrow 1$
		\ELSE
		\STATE $q_{ig(t)}\leftarrow 0$
		\ENDIF
		\ENDFOR
		\ENDFOR
		\ENDWHILE
		\ENDFOR
	\end{algorithmic}
\end{algorithm}

We summarize the overall algorithm for the optimization problem \Cref{matirx8} in Algorithm 1. Note that we start training the classifier using only those reliable instances with high reliability. As the iterations proceed and the training error becomes smaller, more instances will satisfy the constraint \Cref{matirx12} to minimize the loss function of optimization problem \Cref{matirx8}. 
As a result, more and more instances will be selected into the reliable instance set to train the classifier. 
The computational complexity of the first step which updates the variables $\mathbf{w},b$ is $\mathcal{O}(pf^2)$, where $f=\sum_i^n\sum_j^{m_i}q_{ij}$ refers to the number of reliable instances used for training. 
During the second step which selects the reliable instances, the computational complexity of calculating the loss is $\mathcal{O}(nm_ip)$ and the sorting process costs $\mathcal{O}(nm_ilogm_i)$; Moreover, due to the inequality $p>>logm_i$ usually holds in practice, the computational complexity of this step turns to $\mathcal{O}(nm_ip)$ for $i=1,2,\cdots,n$. 
In summary, the computational complexity is $\mathcal{O}\left(pf^2+\mathcal{O}(nm_ip)\right)$ for each iteration of Algorithm 1.


\section{Experiments}

In this section, we conduct thorough experimental evaluations of the proposed framework. Firstly, we compare the proposed algorithm against state-of-the-art alternative baselines. We then compare it against state-of-the-art models using a single feature. After that, we compare it against state-of-the-art systems; finally, we conduct an ablation study to demonstrate the benefit of each component in the proposed algorithm.

\subsection{Experimental setup}

\textbf{Datasets:} Following existing works on Multimedia Event Detection, we evaluate the proposed algorithm on two real-world event detection datasets. These datasets have been compiled by the National Institute of Standard and Technology (NIST) for the TRECVID Multimedia Event Detection competition. To the best of our knowledge, these datasets are the largest public datasets for complex event detection.

\begin{figure}[t]
	\centering
	\includegraphics[width=0.5\textwidth]{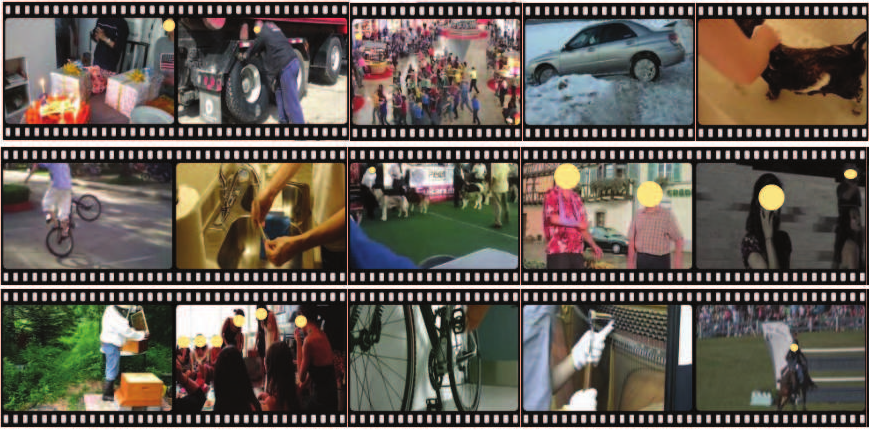}
	\caption{Exemplars from the TRECVID MEDTest 2014 and MEDTest 2013 datasets.}
\end{figure}

\begin{itemize}
	\item MEDTest14~\citep{MEDTest14}: The TRECVID MEDTest 2014 dataset has 100 positive training examples for each event, along with about 5,000 negative samples. There are approximately 23,000 testing videos. This dataset contains events E021 to E040. Some example events are \textit{grooming an animal}, \textit{changing a vehicle tire}, \textit{etc}. Please refer to \citep{MEDTest14, MEDTest13} for a complete list of event names and descriptions.
	\item MEDTest13~\citep{MEDTest13}: Similar to MEDTest14, there are 100 positive training examples for each event, together with about 5,000 negative samples in the MEDTest13 dataset. There are also about 23,000 testing videos. It contains events E006 to E015 and E012 to E030. A complete list of event names and descriptions is provided in \citep{MEDTest13}.
\end{itemize}

\begin{table}[!ht]
	\setlength{\tabcolsep}{9pt}
	\centering
	\caption{Evaluating the performance of the proposed algorithm against alternative baselines. mAP is used as an evaluation metric. The performance is reported in percentages. Larger value indicates better performance.}
	\begin{tabular}{@{}lllcll@{}}
		\toprule
		& \multicolumn{2}{c}{MED14} & \phantom{aaa} & \multicolumn{2}{c}{MED13} \\
		\cmidrule{2-3} \cmidrule{5-6}
		& 100Ex & 10Ex && 100Ex & 10Ex \\
		\midrule
		SVM & 22.8 & 18.3 && 26.9 & 20.7 \\
		RR & 23.6  & 18.8 && 27.5  &  21.2 \\
		Sparse MIL & 19.8 & 14.3 && 23.3 & 16.4 \\
		SIL-SVM & 22.2 & 18.5 && 24.0 & 19.8 \\
		SMIL-TopK & 36.9 & 26.2 && 40.5 & 26.9 \\
		MIL-SRI & 38.6 & 28.4 && 43.1 & 28.7 \\
		Ours  & \textbf{43.2} & \textbf{31.8} && \textbf{48.7} & \textbf{33.9} \\
		\bottomrule 
	\end{tabular}
	\label{tab:alternative_baseline}
\end{table}

\textbf{Setting:} For all experiments, we strictly follow the \textit{100Ex evaluation procedure} outlined in \citep{MEDTest13,MEDTest14}. Following the rules specified in the event kits, we \textbf{separately detect each event}, resulting in 20 individual tasks for each dataset. In other words, event detection is a binary classification task. We use the official split released by the NIST. For each event in the dataset, we have 100 positive training samples, and about 5,000 negative samples. Once the model has been trained, we evaluate it on the testing videos. In this paper, we consider both the \textit{100Ex} and \textit{10Ex} settings provided by the NIST.

\textbf{Feature Extraction:} We first segment each video into multiple shots using the color histogram difference as the indication of the shot boundary. In line with existing works on event detection, we simply choose the center frame from each shot, resize it to $224\times 224$, and extract features from the $fc6$ layer of VGG16 \citep{VGG16}. We run a state-of-the-art video-to-text model on each segment, then generate a description for each segment \citep{Venugopalan2015Sequence}.

\textbf{Evaluation Metric:} Following the NIST standard, we evaluate the event detection performance using mean Average Precision (mAP). Average Precision, which has been widely used in the area of information retrieval, is a single-valued metric approximating the area under the precision-recall curve; mAP is the mean of AP over all event classes.

\begin{table}[t]
\centering
	\setlength{\tabcolsep}{2pt}
	\caption{Performance comparison against state-of-the-art alternatives that use a \textbf{single} type of feature on the TRECVID MEDTest 2014 and MEDTest 2013 datasets. mAP is used as an evaluation metric. Performance is reported in percentages. Larger value indicates better performance.}
	\label{tab:single_feature}		
	\begin{tabular}{@{}cccccc@{}}
		\toprule
		& \multicolumn{2}{c}{MED14} & \phantom{a} & \multicolumn{2}{c}{MED13} 
		\\
		\cmidrule{2-3} \cmidrule{5-6} 	
		& 100Ex & 10Ex && 100Ex & 10Ex \\
		\midrule
		LTS \citep{tang2012learning} & 27.5 & 16.8 &
		& 34.6 & 18.2 
		\\
		SED \citep{Laietal14}  & 29.6 & 18.4 &
		& 36.2 & 20.1 
		\\
		DP \citep{Lietal13} & 28.8  & 17.6 &
		& 35.3  & 19.5 
		\\
		STN \citep{KarpathyTSLSF14} & 30.4  & 19.8 &
		& 37.1  & 20.4 
		\\
		C3D \citep{tran2015learning} & 31.4  & 20.5 &
		& 36.9  & 22.2 
		\\
		MIFS \citep{lan2015beyond} & 29.0  &  14.9 & 
		& 36.3  &  19.3 
		\\
		CNN-Exp \citep{zha2015exploiting} & 29.7  &  -- & 
		& --  &  -- 
		\\
		CNN + VLAD \citep{xu2015discriminative} & 35.7  &  23.2 & 
		& 40.3  &  25.6 
		\\
		NI-SVM \citep{chang2017semantic}  & 34.4  &  26.1 & 
		& 39.2  &  26.8
		\\
		MIL-SRI \citep{fan2017complex} & 38.6 & 28.4 & & 43.1 & 28.7 \\
		Ours & \textbf{43.2} & \textbf{31.8} & & \textbf{48.7} & \textbf{33.9}
		\\
		\bottomrule 
	\end{tabular}
\end{table}

\subsection{Comparison against alternative baselines}

In this section, we compare the performance of the proposed algorithm against state-of-the-art alternative baselines. More specifically, we conduct comparisons against the following:

\begin{itemize}
	\item \textbf{Support Vector Machine (SVM) and Ridge Regression (RR):} SVM and RR are the commonly used classifiers in the TRECVID Multimedia Event Detection (MED) competition among the top ranked teams and existing technical reports.
	\item \textbf{Sparse Multi-Instance Learning (Sparse MIL):} We calculate the central point for each bag via the operation of average-pooling on all instances, based on which we aim to train a bag-level classifier.
	\item \textbf{SIL-SVM:} We first assign instances' labels as the corresponding bags' labels, then train an instance-level classifier using this information.
	\item \textbf{SMIL-TopK:} We first select the most confident $k$ instances in each bag, and then train an instance-level classifier based on the selected instances. The optimal parameter $k$ is tuned via cross-validation.
	\item \textbf{Multi-Instance Learning by Selecting Reliable Instances (MIL-SRI):} MIL-SRI aims to adaptively select reliable instances and does not require the inference of instance labels. We tune the parameters in the range of $\{10^{-4}, 10^{-3}, 10^{-2}, 10^{-1}, 1, 10\}$ and select the best one using cross-validation.
\end{itemize}

Experimental results are reported in \Cref{tab:alternative_baseline}; from these results, we can make the following observations. Firstly, the traditional SVM and RR can obtain promising results in both settings on the datasets of interest. Secondly, Sparse MIL and SIL-SVM achieve slightly worse performance on these datasets, and this is mainly because they did not differentiate the instances. Thirdly, we observe that SMIL-TopK and MIL-SRI significantly improve the performance of event detection on the used datasets, which indicates the benefits of exploiting reliable shots. Lastly, the algorithm proposed in this paper outperforms the other baselines by a large margin. This demonstrates the benefits of jointly exploring the semantic and visual feature for reliability learning.

\begin{figure*}[t]
	\centering
	\includegraphics[scale=0.4]{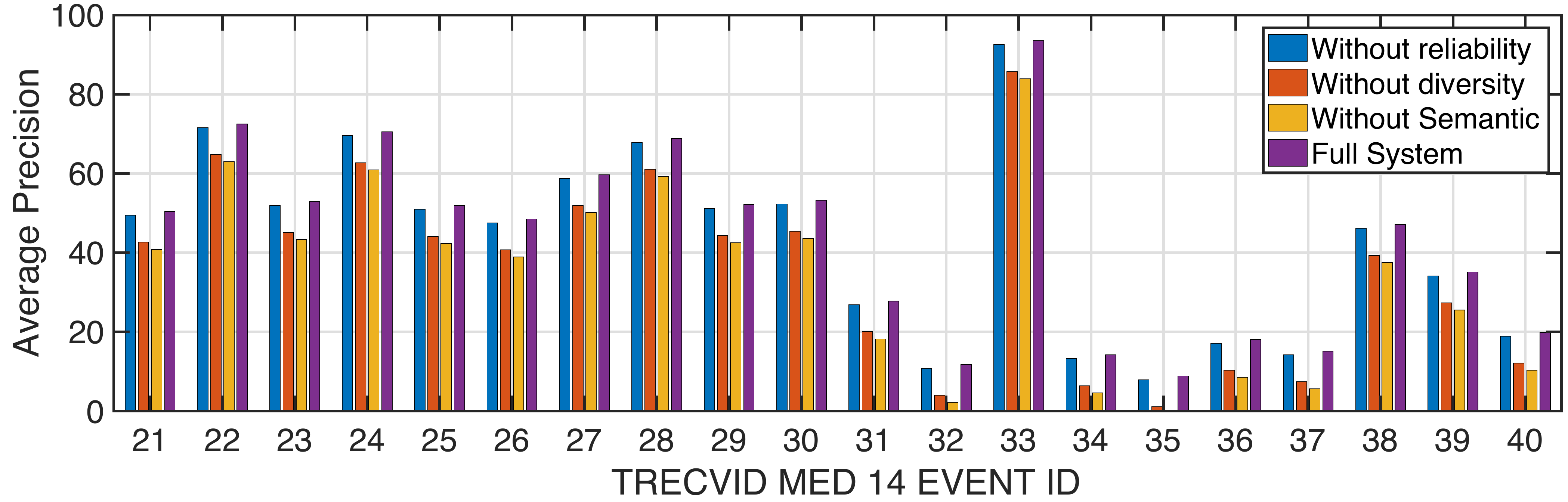}
	\caption{We compare the results of different versions of the proposed method, including (a) without reliability; (b) without diversity; (c) without semantic; (d) full system.}
	\label{fig:ablation}
\end{figure*}

\begin{table}[t]
	\setlength{\tabcolsep}{0.1pt}
	\centering
	\caption{Comparison of the proposed model against state-of-the-art systems on the TRECVID MEDTest 2014 and MEDTest 2013 datasets. Performance is reported in percentages. Larger value indicates better performance.}
	\begin{tabular}{@{}cccccc@{}}
		\toprule
		& \multicolumn{2}{c}{MED14} & \phantom{aaa} & \multicolumn{2}{c}{MED13} \\
		\cmidrule{2-3} \cmidrule{5-6}
		& 100Ex & 10Ex && 100Ex & 10Ex \\
		\midrule
		C3D \citep{tran2015learning} + IDT & 33.6 & 22.1 && 39.5 & 26.7 \\
		CNN-Exp \citep{zha2015exploiting}  & 38.7 & -- && -- & -- \\
		CNN + VLAD \citep{xu2015discriminative}  & 36.8 & 24.5 && 44.6 & 29.8 \\
		NISVM + IDT \citep{chang2017semantic} & 38.1 & 27.2 && 46.3 & 31.5 \\
		MIL-SRI + IDT \citep{fan2017complex} & 41.5 & 29.6 &  & 49.7 & 34.6 \\
		Ours + IDT & \textbf{44.9} & \textbf{33.5} & & \textbf{50.3} & \textbf{35.2} \\
		\bottomrule 
	\end{tabular}
	\label{tab:multiple_feature}
\end{table}

\subsection{Comparison under a single feature}

To further validate the effectiveness of the proposed model, we compare the proposed algorithm with state-of-the-art alternatives that use a \textbf{single} type of feature. The experimental results are presented in \Cref{tab:single_feature}. Note that, whenever possible, we directly quote the result from the original reference; in cases where this result was not directly available, we requested the code from the authors and ran the experiments ourselves. 

From the experimental results in \Cref{tab:single_feature}, we can clearly see that the proposed algorithm outperforms the other models with a single feature by a large margin. For example, on the TRECVID MEDTest 2014 dataset, the proposed method achieves 43.2\% mAP, which outperforms the second best model, MIL-SRI by 4.6\% in the 100Ex setting, while also outperforming MIL-SRI by 3.4\% in the 10Ex setting. This improvement is significant in the TRECVID Multimedia Event Detection competition, since event detection is a very challenging task.

\subsection{Comparison with state-of-the-art systems}

In the TRECVID Multimedia Event Detection competition, the top teams explore different ways to combine \textbf{multiple} different types of features. 
Accordingly, in this section, we also compare our method with state-of-the-art systems in the literature. In the last few years, Improved Dense Trajectories (IDT) \citep{Wang2014Action} have significantly outperformed the other features for the multimedia event detection competition; hence, to facilitate fair comparison, we also combine the prediction result of our method with that of IDT.
The experimental results are presented in \Cref{tab:multiple_feature}. From these results, we can see that with the proposed model, a simple combination with IDT can significantly outperform state-of-the-art systems. This further demonstrates the effectiveness of the proposed model.

\subsection{Ablation study}
In this section, additional experiments are conducted to confirm the effectiveness of different terms in the full system. In more detail, we compare the full system against (a) full system without reliability; (b) full system without diversity; and (c) full system without semantic. The detailed performance on an individual event is plotted in \Cref{fig:ablation}. From the results, we can see that the full system consistently outperforms the other three alternatives; this confirms the effectiveness of all three functions. We can also observe that dropping the semantic part results in the most significant decline in performance. This phenomenon demonstrates the importance of incorporating semantic information in order to learn the reliability for each instance.

\section{Conclusion}
In this paper, we propose a novel approach to event detection in the framework of multi-instance learning, which learn the reliability of each instance by jointly exploiting both visual and semantic information simultaneously. To improve the robustness of the classifier, we begin the training process using high-reliability instances and gradually added in instances with relatively low reliability over time. This strategy alleviates the negative influence of irrelevant and ambiguous segments in the training process. The proposed algorithm was evaluated on two large-scale challenging datasets, and achieved very promising results. A possible direction for future work on event detection may lie in exploiting contrastive learning of visual and semantic information to extract better representations for multi-instance learning.

\section*{Acknowledgement}
This work was supported by National Nature Science Foundation of China (No. 61872287 and No. 61973162), Innovative Research Group of the National Natural Science Foundation of China (No. 61721002), Innovation Research Team of Ministry of Education (IRT\_17R86), Project of China Knowledge Center for Engineering Science and Technology, NSF of Jiangsu Province (No. BZ2021013), the Fundamental Research Funds for the Central Universities (Nos: 30920032202, 30921013114), and Australian Research Council (ARC) Discovery Early Career Researcher Award (DECRA) under DE190100626.

\bibliographystyle{model2-names}
\bibliography{refs}

\begin{thebibliography}{91}
\expandafter\ifx\csname natexlab\endcsname\relax\def\natexlab#1{#1}\fi
\providecommand{\url}[1]{\texttt{#1}}
\providecommand{\href}[2]{#2}
\providecommand{\path}[1]{#1}
\providecommand{\DOIprefix}{doi:}
\providecommand{\ArXivprefix}{arXiv:}
\providecommand{\URLprefix}{URL: }
\providecommand{\Pubmedprefix}{pmid:}
\providecommand{\doi}[1]{\href{http://dx.doi.org/#1}{\path{#1}}}
\providecommand{\Pubmed}[1]{\href{pmid:#1}{\path{#1}}}
\providecommand{\bibinfo}[2]{#2}
\ifx\xfnm\relax \def\xfnm[#1]{\unskip,\space#1}\fi
\bibitem[{Bengio et~al.(2009)Bengio, Louradour, Collobert and
  Weston}]{bengio2009curriculum}
\bibinfo{author}{Bengio, Y.}, \bibinfo{author}{Louradour, J.},
  \bibinfo{author}{Collobert, R.}, \bibinfo{author}{Weston, J.},
  \bibinfo{year}{2009}.
\newblock \bibinfo{title}{Curriculum learning}, in:
  \bibinfo{booktitle}{Proceedings of the 26th International Conference on
  Machine Learning}, pp. \bibinfo{pages}{41--48}.
\bibitem[{Bunescu and Mooney(2007)}]{bunescu2007multiple}
\bibinfo{author}{Bunescu, R.C.}, \bibinfo{author}{Mooney, R.J.},
  \bibinfo{year}{2007}.
\newblock \bibinfo{title}{Multiple instance learning for sparse positive bags},
  in: \bibinfo{booktitle}{Proceedings of the 24th International Conference on
  Machine Learning}, pp. \bibinfo{pages}{105--112}.
\bibitem[{Chakraborty et~al.(2013)Chakraborty, Gonzalez and
  Roca}]{chakraborty2013large}
\bibinfo{author}{Chakraborty, B.}, \bibinfo{author}{Gonzalez, J.},
  \bibinfo{author}{Roca, F.X.}, \bibinfo{year}{2013}.
\newblock \bibinfo{title}{Large scale continuous visual event recognition using
  max-margin hough transformation framework}.
\newblock \bibinfo{journal}{Computer Vision and Image Understanding}
  \bibinfo{volume}{117}, \bibinfo{pages}{1356--1368}.
\bibitem[{Chang et~al.(2017a)Chang, Ma, Lin, Yang and Hauptmann}]{ChangMLYH17}
\bibinfo{author}{Chang, X.}, \bibinfo{author}{Ma, Z.}, \bibinfo{author}{Lin,
  M.}, \bibinfo{author}{Yang, Y.}, \bibinfo{author}{Hauptmann, A.G.},
  \bibinfo{year}{2017}a.
\newblock \bibinfo{title}{Feature interaction augmented sparse learning for
  fast kinect motion detection}.
\newblock \bibinfo{journal}{{IEEE} Trans. Image Process.} \bibinfo{volume}{26},
  \bibinfo{pages}{3911--3920}.
\bibitem[{Chang et~al.(2016a)Chang, Ma, Yang, Zeng and Hauptmann}]{chang2016bi}
\bibinfo{author}{Chang, X.}, \bibinfo{author}{Ma, Z.}, \bibinfo{author}{Yang,
  Y.}, \bibinfo{author}{Zeng, Z.}, \bibinfo{author}{Hauptmann, A.G.},
  \bibinfo{year}{2016}a.
\newblock \bibinfo{title}{Bi-level semantic representation analysis for
  multimedia event detection}.
\newblock \bibinfo{journal}{IEEE Transactions on Cybernetics}
  \bibinfo{volume}{47}, \bibinfo{pages}{1180--1197}.
\bibitem[{Chang and Yang(2017)}]{ChangY17}
\bibinfo{author}{Chang, X.}, \bibinfo{author}{Yang, Y.}, \bibinfo{year}{2017}.
\newblock \bibinfo{title}{Semisupervised feature analysis by mining
  correlations among multiple tasks}.
\newblock \bibinfo{journal}{{IEEE} Trans. Neural Networks Learn. Syst.}
  \bibinfo{volume}{28}, \bibinfo{pages}{2294--2305}.
\bibitem[{Chang et~al.(2016b)Chang, Yu, Yang and Xing}]{ChangYYX16}
\bibinfo{author}{Chang, X.}, \bibinfo{author}{Yu, Y.}, \bibinfo{author}{Yang,
  Y.}, \bibinfo{author}{Xing, E.P.}, \bibinfo{year}{2016}b.
\newblock \bibinfo{title}{They are not equally reliable: Semantic event search
  using differentiated concept classifiers}, in: \bibinfo{booktitle}{2016
  {IEEE} Conference on Computer Vision and Pattern Recognition, {CVPR} 2016,
  Las Vegas, NV, USA, June 27-30, 2016}, pp. \bibinfo{pages}{1884--1893}.
\bibitem[{Chang et~al.(2017b)Chang, Yu, Yang and Xing}]{chang2017semantic}
\bibinfo{author}{Chang, X.}, \bibinfo{author}{Yu, Y.}, \bibinfo{author}{Yang,
  Y.}, \bibinfo{author}{Xing, E.P.}, \bibinfo{year}{2017}b.
\newblock \bibinfo{title}{Semantic pooling for complex event analysis in
  untrimmed videos}.
\newblock \bibinfo{journal}{IEEE Transactions on Pattern Analysis and Machine
  Intelligence} \bibinfo{volume}{39}, \bibinfo{pages}{1617--1632}.
\bibitem[{Chen et~al.(2020)Chen, Yao, Zhang, Wang, Chang and Nie}]{ChenYZWCN20}
\bibinfo{author}{Chen, K.}, \bibinfo{author}{Yao, L.}, \bibinfo{author}{Zhang,
  D.}, \bibinfo{author}{Wang, X.}, \bibinfo{author}{Chang, X.},
  \bibinfo{author}{Nie, F.}, \bibinfo{year}{2020}.
\newblock \bibinfo{title}{A semisupervised recurrent convolutional attention
  model for human activity recognition}.
\newblock \bibinfo{journal}{{IEEE} Trans. Neural Networks Learn. Syst.}
  \bibinfo{volume}{31}, \bibinfo{pages}{1747--1756}.
\bibitem[{Chen et~al.(2006)Chen, Bi and Wang}]{chen2006miles}
\bibinfo{author}{Chen, Y.}, \bibinfo{author}{Bi, J.}, \bibinfo{author}{Wang,
  J.Z.}, \bibinfo{year}{2006}.
\newblock \bibinfo{title}{Miles: Multiple-instance learning via embedded
  instance selection}.
\newblock \bibinfo{journal}{IEEE Transactions on Pattern Analysis and Machine
  Intelligence} \bibinfo{volume}{28}, \bibinfo{pages}{1931--1947}.
\bibitem[{Chen and Wang(2004)}]{chen2004image}
\bibinfo{author}{Chen, Y.}, \bibinfo{author}{Wang, J.Z.}, \bibinfo{year}{2004}.
\newblock \bibinfo{title}{Image categorization by learning and reasoning with
  regions}.
\newblock \bibinfo{journal}{Journal of Machine Learning Research}
  \bibinfo{volume}{5}, \bibinfo{pages}{913--939}.
\bibitem[{Cheng et~al.(2019)Cheng, Chang, Zhu, Kanjirathinkal and
  Kankanhalli}]{ChengCZKK19}
\bibinfo{author}{Cheng, Z.}, \bibinfo{author}{Chang, X.}, \bibinfo{author}{Zhu,
  L.}, \bibinfo{author}{Kanjirathinkal, R.C.}, \bibinfo{author}{Kankanhalli,
  M.S.}, \bibinfo{year}{2019}.
\newblock \bibinfo{title}{{MMALFM:} explainable recommendation by leveraging
  reviews and images}.
\newblock \bibinfo{journal}{{ACM} Trans. Inf. Syst.} \bibinfo{volume}{37},
  \bibinfo{pages}{16:1--16:28}.
\bibitem[{Cheny et~al.(2019)Cheny, Fuy, Zhang, Jiang, Xue and
  Sigal}]{cheny2019multi}
\bibinfo{author}{Cheny, Z.}, \bibinfo{author}{Fuy, Y.}, \bibinfo{author}{Zhang,
  Y.}, \bibinfo{author}{Jiang, Y.G.}, \bibinfo{author}{Xue, X.},
  \bibinfo{author}{Sigal, L.}, \bibinfo{year}{2019}.
\newblock \bibinfo{title}{Multi-level semantic feature augmentation for
  one-shot learning}.
\newblock \bibinfo{journal}{IEEE Transactions on Image Processing}
  \bibinfo{volume}{28}, \bibinfo{pages}{4594--4605}.
\bibitem[{Dietterich et~al.(1997)Dietterich, Lathrop and
  Lozano-P{\'e}rez}]{dietterich1997solving}
\bibinfo{author}{Dietterich, T.G.}, \bibinfo{author}{Lathrop, R.H.},
  \bibinfo{author}{Lozano-P{\'e}rez, T.}, \bibinfo{year}{1997}.
\newblock \bibinfo{title}{Solving the multiple instance problem with
  axis-parallel rectangles}.
\newblock \bibinfo{journal}{Artificial Intelligence} \bibinfo{volume}{89},
  \bibinfo{pages}{31--71}.
\bibitem[{Fan et~al.(2017)Fan, Chang, Cheng, Yang, Xu and
  Hauptmann}]{fan2017complex}
\bibinfo{author}{Fan, H.}, \bibinfo{author}{Chang, X.}, \bibinfo{author}{Cheng,
  D.}, \bibinfo{author}{Yang, Y.}, \bibinfo{author}{Xu, D.},
  \bibinfo{author}{Hauptmann, A.G.}, \bibinfo{year}{2017}.
\newblock \bibinfo{title}{Complex event detection by identifying reliable shots
  from untrimmed videos}, in: \bibinfo{booktitle}{Proceedings of the IEEE
  International Conference on Computer Vision}, pp. \bibinfo{pages}{736--744}.
\bibitem[{G{\"a}rtner et~al.(2002)G{\"a}rtner, Flach, Kowalczyk and
  Smola}]{gartner2002multi}
\bibinfo{author}{G{\"a}rtner, T.}, \bibinfo{author}{Flach, P.A.},
  \bibinfo{author}{Kowalczyk, A.}, \bibinfo{author}{Smola, A.J.},
  \bibinfo{year}{2002}.
\newblock \bibinfo{title}{Multi-instance kernels}, in:
  \bibinfo{booktitle}{Proceedings of the International Conference on Machine
  Learning}, pp. \bibinfo{pages}{179--186}.
\bibitem[{Ghasedi et~al.(2019)Ghasedi, Wang, Deng and
  Huang}]{ghasedi2019balanced}
\bibinfo{author}{Ghasedi, K.}, \bibinfo{author}{Wang, X.},
  \bibinfo{author}{Deng, C.}, \bibinfo{author}{Huang, H.},
  \bibinfo{year}{2019}.
\newblock \bibinfo{title}{Balanced self-paced learning for generative
  adversarial clustering network}, in: \bibinfo{booktitle}{Proceedings of the
  IEEE Conference on Computer Vision and Pattern Recognition}, pp.
  \bibinfo{pages}{4391--4400}.
\bibitem[{Habibian and Snoek(2014)}]{habibian2014recommendations}
\bibinfo{author}{Habibian, A.}, \bibinfo{author}{Snoek, C.G.},
  \bibinfo{year}{2014}.
\newblock \bibinfo{title}{Recommendations for recognizing video events by
  concept vocabularies}.
\newblock \bibinfo{journal}{Computer Vision and Image Understanding}
  \bibinfo{volume}{124}, \bibinfo{pages}{110--122}.
\bibitem[{Huang et~al.(2019)Huang, Shi, Zhang, Wu and
  Chawla}]{huang2019similarity}
\bibinfo{author}{Huang, C.}, \bibinfo{author}{Shi, B.}, \bibinfo{author}{Zhang,
  X.}, \bibinfo{author}{Wu, X.}, \bibinfo{author}{Chawla, N.V.},
  \bibinfo{year}{2019}.
\newblock \bibinfo{title}{Similarity-aware network embedding with self-paced
  learning}, in: \bibinfo{booktitle}{Proceedings of the 28th ACM International
  Conference on Information and Knowledge Management}, pp.
  \bibinfo{pages}{2113--2116}.
\bibitem[{Jiang et~al.(2012)Jiang, Hauptmann and Xiang}]{jiang2012leveraging}
\bibinfo{author}{Jiang, L.}, \bibinfo{author}{Hauptmann, A.G.},
  \bibinfo{author}{Xiang, G.}, \bibinfo{year}{2012}.
\newblock \bibinfo{title}{Leveraging high-level and low-level features for
  multimedia event detection}, in: \bibinfo{booktitle}{Proceedings of the 20th
  ACM International Conference on Multimedia}, pp. \bibinfo{pages}{449--458}.
\bibitem[{Jiang et~al.(2014a)Jiang, Meng, Mitamura and
  Hauptmann}]{jiang2014easy}
\bibinfo{author}{Jiang, L.}, \bibinfo{author}{Meng, D.},
  \bibinfo{author}{Mitamura, T.}, \bibinfo{author}{Hauptmann, A.G.},
  \bibinfo{year}{2014}a.
\newblock \bibinfo{title}{Easy samples first: Self-paced reranking for
  zero-example multimedia search}, in: \bibinfo{booktitle}{Proceedings of the
  22nd ACM International Conference on Multimedia}, pp.
  \bibinfo{pages}{547--556}.
\bibitem[{Jiang et~al.(2014b)Jiang, Meng, Yu, Lan, Shan and
  Hauptmann}]{Jiang2014Self}
\bibinfo{author}{Jiang, L.}, \bibinfo{author}{Meng, D.}, \bibinfo{author}{Yu,
  S.I.}, \bibinfo{author}{Lan, Z.}, \bibinfo{author}{Shan, S.},
  \bibinfo{author}{Hauptmann, A.}, \bibinfo{year}{2014}b.
\newblock \bibinfo{title}{Self-paced learning with diversity}.
\newblock \bibinfo{journal}{Proceedings of the Advances in Neural Information
  Processing Systems} , \bibinfo{pages}{2078--2086}.
\bibitem[{Karpathy et~al.(2014a)Karpathy, Toderici, Shetty, Leung, Sukthankar
  and Fei-Fei}]{Karpathy2014Large}
\bibinfo{author}{Karpathy, A.}, \bibinfo{author}{Toderici, G.},
  \bibinfo{author}{Shetty, S.}, \bibinfo{author}{Leung, T.},
  \bibinfo{author}{Sukthankar, R.}, \bibinfo{author}{Fei-Fei, L.},
  \bibinfo{year}{2014}a.
\newblock \bibinfo{title}{Large-scale video classification with convolutional
  neural networks}, in: \bibinfo{booktitle}{Proceedings of the IEEE Conference
  on Computer Vision and Pattern Recognition}, pp. \bibinfo{pages}{1725--1732}.
\bibitem[{Karpathy et~al.(2014b)Karpathy, Toderici, Shetty, Leung, Sukthankar
  and Li}]{KarpathyTSLSF14}
\bibinfo{author}{Karpathy, A.}, \bibinfo{author}{Toderici, G.},
  \bibinfo{author}{Shetty, S.}, \bibinfo{author}{Leung, T.},
  \bibinfo{author}{Sukthankar, R.}, \bibinfo{author}{Li, F.},
  \bibinfo{year}{2014}b.
\newblock \bibinfo{title}{Large-scale video classification with convolutional
  neural networks}, in: \bibinfo{booktitle}{Proceedings of the IEEE Conference
  on Computer Vision and Pattern Recognition}, pp. \bibinfo{pages}{1725--1732}.
\bibitem[{Krizhevsky et~al.(2012)Krizhevsky, Sutskever and
  Hinton}]{krizhevsky2012imagenet}
\bibinfo{author}{Krizhevsky, A.}, \bibinfo{author}{Sutskever, I.},
  \bibinfo{author}{Hinton, G.E.}, \bibinfo{year}{2012}.
\newblock \bibinfo{title}{Imagenet classification with deep convolutional
  neural networks}, in: \bibinfo{booktitle}{Proceedings of the Advances in
  Neural Information Processing Systems}, pp. \bibinfo{pages}{1097--1105}.
\bibitem[{Kumar et~al.(2010)Kumar, Packer and Koller}]{kumar2010self-paced}
\bibinfo{author}{Kumar, M.P.}, \bibinfo{author}{Packer, B.},
  \bibinfo{author}{Koller, D.}, \bibinfo{year}{2010}.
\newblock \bibinfo{title}{Self-paced learning for latent variable models}, in:
  \bibinfo{booktitle}{Proceedings of the Advances in Neural Information
  Processing Systems}, pp. \bibinfo{pages}{1189--1197}.
\bibitem[{Lai et~al.(2014a)Lai, Liu, Chen and Chang}]{Laietal14}
\bibinfo{author}{Lai, K.T.}, \bibinfo{author}{Liu, D.}, \bibinfo{author}{Chen,
  M.S.}, \bibinfo{author}{Chang, S.F.}, \bibinfo{year}{2014}a.
\newblock \bibinfo{title}{Recognizing complex events in videos by learning key
  static-dynamic evidences}, in: \bibinfo{booktitle}{Proceedings of the
  European Conference on Computer Vision}, pp. \bibinfo{pages}{675--688}.
\bibitem[{Lai et~al.(2014b)Lai, Yu, Chen and Chang}]{lai2014video}
\bibinfo{author}{Lai, K.T.}, \bibinfo{author}{Yu, F.X.}, \bibinfo{author}{Chen,
  M.S.}, \bibinfo{author}{Chang, S.F.}, \bibinfo{year}{2014}b.
\newblock \bibinfo{title}{Video event detection by inferring temporal instance
  labels}, in: \bibinfo{booktitle}{Proceedings of the IEEE Conference on
  Computer Vision and Pattern Recognition}, pp. \bibinfo{pages}{2243--2250}.
\bibitem[{Lan et~al.(2015)Lan, Lin, Li, Hauptmann and Raj}]{lan2015beyond}
\bibinfo{author}{Lan, Z.}, \bibinfo{author}{Lin, M.}, \bibinfo{author}{Li, X.},
  \bibinfo{author}{Hauptmann, A.G.}, \bibinfo{author}{Raj, B.},
  \bibinfo{year}{2015}.
\newblock \bibinfo{title}{Beyond gaussian pyramid: Multi-skip feature stacking
  for action recognition}, in: \bibinfo{booktitle}{Proceedings of the IEEE
  Conference on Computer Vision and Pattern Recognition}, pp.
  \bibinfo{pages}{204--212}.
\bibitem[{Laptev(2005)}]{laptev2005space}
\bibinfo{author}{Laptev, I.}, \bibinfo{year}{2005}.
\newblock \bibinfo{title}{On space-time interest points}.
\newblock \bibinfo{journal}{International Journal of Computer Vision}
  \bibinfo{volume}{64}, \bibinfo{pages}{107--123}.
\bibitem[{Li et~al.(2016)Li, Gong, Meng and Miao}]{li2016multi-objective}
\bibinfo{author}{Li, H.}, \bibinfo{author}{Gong, M.}, \bibinfo{author}{Meng,
  D.}, \bibinfo{author}{Miao, Q.}, \bibinfo{year}{2016}.
\newblock \bibinfo{title}{Multi-objective self-paced learning}, in:
  \bibinfo{booktitle}{Proceedings of the 30th AAAI Conference on Artificial
  Intelligence}, pp. \bibinfo{pages}{1802--1808}.
\bibitem[{Li et~al.(2011)Li, Duan, Xu and Tsang}]{li2011text}
\bibinfo{author}{Li, W.}, \bibinfo{author}{Duan, L.}, \bibinfo{author}{Xu, D.},
  \bibinfo{author}{Tsang, I.W.H.}, \bibinfo{year}{2011}.
\newblock \bibinfo{title}{Text-based image retrieval using progressive
  multi-instance learning}, in: \bibinfo{booktitle}{Proceedings of the IEEE
  International Conference on Computer Vision}, pp.
  \bibinfo{pages}{2049--2055}.
\bibitem[{Li and Vasconcelos(2015)}]{li2015multiple}
\bibinfo{author}{Li, W.}, \bibinfo{author}{Vasconcelos, N.},
  \bibinfo{year}{2015}.
\newblock \bibinfo{title}{Multiple instance learning for soft bags via top
  instances}, in: \bibinfo{booktitle}{Proceedings of the IEEE Conference on
  Computer Vision and Pattern Recognition}, pp. \bibinfo{pages}{4277--4285}.
\bibitem[{Li et~al.(2013)Li, Yu, Divakaran and Vasconcelos}]{Lietal13}
\bibinfo{author}{Li, W.}, \bibinfo{author}{Yu, Q.}, \bibinfo{author}{Divakaran,
  A.}, \bibinfo{author}{Vasconcelos, N.}, \bibinfo{year}{2013}.
\newblock \bibinfo{title}{Dynamic pooling for complex event recognition}, in:
  \bibinfo{booktitle}{Proceedings of the IEEE International Conference on
  Computer Vision}, pp. \bibinfo{pages}{2728--2735}.
\bibitem[{Li et~al.(2018a)Li, Nie, Chang, Nie, Zhang and Yang}]{LiNCNZY18}
\bibinfo{author}{Li, Z.}, \bibinfo{author}{Nie, F.}, \bibinfo{author}{Chang,
  X.}, \bibinfo{author}{Nie, L.}, \bibinfo{author}{Zhang, H.},
  \bibinfo{author}{Yang, Y.}, \bibinfo{year}{2018}a.
\newblock \bibinfo{title}{Rank-constrained spectral clustering with flexible
  embedding}.
\newblock \bibinfo{journal}{{IEEE} Trans. Neural Networks Learn. Syst.}
  \bibinfo{volume}{29}, \bibinfo{pages}{6073--6082}.
\bibitem[{Li et~al.(2018b)Li, Nie, Chang, Yang, Zhang and Sebe}]{LiNCYZS18}
\bibinfo{author}{Li, Z.}, \bibinfo{author}{Nie, F.}, \bibinfo{author}{Chang,
  X.}, \bibinfo{author}{Yang, Y.}, \bibinfo{author}{Zhang, C.},
  \bibinfo{author}{Sebe, N.}, \bibinfo{year}{2018}b.
\newblock \bibinfo{title}{Dynamic affinity graph construction for spectral
  clustering using multiple features}.
\newblock \bibinfo{journal}{{IEEE} Trans. Neural Networks Learn. Syst.}
  \bibinfo{volume}{29}, \bibinfo{pages}{6323--6332}.
\bibitem[{Li et~al.(2019)Li, Yao, Chang, Zhan, Sun and Zhang}]{li2019zero-shot}
\bibinfo{author}{Li, Z.}, \bibinfo{author}{Yao, L.}, \bibinfo{author}{Chang,
  X.}, \bibinfo{author}{Zhan, K.}, \bibinfo{author}{Sun, J.},
  \bibinfo{author}{Zhang, H.}, \bibinfo{year}{2019}.
\newblock \bibinfo{title}{Zero-shot event detection via event-adaptive concept
  relevance mining}.
\newblock \bibinfo{journal}{Pattern Recognition} \bibinfo{volume}{88},
  \bibinfo{pages}{595--603}.
\bibitem[{Liu et~al.(2012)Liu, Wu and Zhou}]{liu2012key}
\bibinfo{author}{Liu, G.}, \bibinfo{author}{Wu, J.}, \bibinfo{author}{Zhou,
  Z.H.}, \bibinfo{year}{2012}.
\newblock \bibinfo{title}{Key instance detection in multi-instance learning},
  in: \bibinfo{booktitle}{Proceedings of the Asian Conference on Machine
  Learning}, pp. \bibinfo{pages}{253--268}.
\bibitem[{Lowe(2004)}]{lowe2004distinctive}
\bibinfo{author}{Lowe, D.G.}, \bibinfo{year}{2004}.
\newblock \bibinfo{title}{Distinctive image features from scale-invariant
  keypoints}.
\newblock \bibinfo{journal}{International Journal of Computer Vision}
  \bibinfo{volume}{60}, \bibinfo{pages}{91--110}.
\bibitem[{Luo et~al.(2017)Luo, Chang, Li, Nie, Hauptmann and
  Zheng}]{LuoCLNHZ17}
\bibinfo{author}{Luo, M.}, \bibinfo{author}{Chang, X.}, \bibinfo{author}{Li,
  Z.}, \bibinfo{author}{Nie, L.}, \bibinfo{author}{Hauptmann, A.G.},
  \bibinfo{author}{Zheng, Q.}, \bibinfo{year}{2017}.
\newblock \bibinfo{title}{Simple to complex cross-modal learning to rank}.
\newblock \bibinfo{journal}{Comput. Vis. Image Underst.} \bibinfo{volume}{163},
  \bibinfo{pages}{67--77}.
\bibitem[{Luo et~al.(2018)Luo, Nie, Chang, Yang, Hauptmann and
  Zheng}]{LuoNCYHZ18}
\bibinfo{author}{Luo, M.}, \bibinfo{author}{Nie, F.}, \bibinfo{author}{Chang,
  X.}, \bibinfo{author}{Yang, Y.}, \bibinfo{author}{Hauptmann, A.G.},
  \bibinfo{author}{Zheng, Q.}, \bibinfo{year}{2018}.
\newblock \bibinfo{title}{Adaptive unsupervised feature selection with
  structure regularization}.
\newblock \bibinfo{journal}{{IEEE} Trans. Neural Networks Learn. Syst.}
  \bibinfo{volume}{29}, \bibinfo{pages}{944--956}.
\bibitem[{Ma et~al.(2017)Ma, Chang, Yang, Sebe and Hauptmann}]{MaCYSH17}
\bibinfo{author}{Ma, Z.}, \bibinfo{author}{Chang, X.}, \bibinfo{author}{Yang,
  Y.}, \bibinfo{author}{Sebe, N.}, \bibinfo{author}{Hauptmann, A.G.},
  \bibinfo{year}{2017}.
\newblock \bibinfo{title}{The many shades of negativity}.
\newblock \bibinfo{journal}{{IEEE} Trans. Multim.} \bibinfo{volume}{19},
  \bibinfo{pages}{1558--1568}.
\bibitem[{Ma et~al.(2013)Ma, Yang, Xu, Yan, Sebe and Hauptmann}]{ma2013complex}
\bibinfo{author}{Ma, Z.}, \bibinfo{author}{Yang, Y.}, \bibinfo{author}{Xu, Z.},
  \bibinfo{author}{Yan, S.}, \bibinfo{author}{Sebe, N.},
  \bibinfo{author}{Hauptmann, A.G.}, \bibinfo{year}{2013}.
\newblock \bibinfo{title}{Complex event detection via multi-source video
  attributes}, in: \bibinfo{booktitle}{Proceedings of the IEEE conference on
  Computer Vision and Pattern Recognition}, pp. \bibinfo{pages}{2627--2633}.
\bibitem[{Mazloom et~al.(2013)Mazloom, Gavves, van~de Sande and
  Snoek}]{mazloom2013searching}
\bibinfo{author}{Mazloom, M.}, \bibinfo{author}{Gavves, E.},
  \bibinfo{author}{van~de Sande, K.}, \bibinfo{author}{Snoek, C.},
  \bibinfo{year}{2013}.
\newblock \bibinfo{title}{Searching informative concept banks for video event
  detection}, in: \bibinfo{booktitle}{Proceedings of ACM conference on
  International Conference on Multimedia Retrieval}, pp.
  \bibinfo{pages}{255--262}.
\bibitem[{Mikolov et~al.(2013)Mikolov, Sutskever, Chen, Corrado and
  Dean}]{mikolov2013distributed}
\bibinfo{author}{Mikolov, T.}, \bibinfo{author}{Sutskever, I.},
  \bibinfo{author}{Chen, K.}, \bibinfo{author}{Corrado, G.S.},
  \bibinfo{author}{Dean, J.}, \bibinfo{year}{2013}.
\newblock \bibinfo{title}{Distributed representations of words and phrases and
  their compositionality}, in: \bibinfo{booktitle}{Proceedings of the Advances
  in Neural Information Processing Systems}, pp. \bibinfo{pages}{3111--3119}.
\bibitem[{Natarajan et~al.(2012)Natarajan, Wu, Vitaladevuni, Zhuang,
  Tsakalidis, Park, Prasad and Natarajan}]{natarajan2012multimodal}
\bibinfo{author}{Natarajan, P.}, \bibinfo{author}{Wu, S.},
  \bibinfo{author}{Vitaladevuni, S.}, \bibinfo{author}{Zhuang, X.},
  \bibinfo{author}{Tsakalidis, S.}, \bibinfo{author}{Park, U.},
  \bibinfo{author}{Prasad, R.}, \bibinfo{author}{Natarajan, P.},
  \bibinfo{year}{2012}.
\newblock \bibinfo{title}{Multimodal feature fusion for robust event detection
  in web videos}, in: \bibinfo{booktitle}{Proceedings of the IEEE Conference on
  Computer Vision and Pattern Recognition}, pp. \bibinfo{pages}{1298--1305}.
\bibitem[{NIST(2013)}]{MEDTest13}
\bibinfo{author}{NIST}, \bibinfo{year}{2013}.
\newblock \bibinfo{title}{The trecvid med 2013 dataset}.
\newblock \bibinfo{howpublished}{\url{http://nist.gov/itl/iad/mig/med13.cfm}}.
\bibitem[{NIST(2014)}]{MEDTest14}
\bibinfo{author}{NIST}, \bibinfo{year}{2014}.
\newblock \bibinfo{title}{The trecvid med 2014 dataset}.
\newblock \bibinfo{howpublished}{\url{http://nist.gov/itl/iad/mig/med14.cfm}}.
\bibitem[{Oneata et~al.(2013)Oneata, Verbeek and Schmid}]{oneata2013action}
\bibinfo{author}{Oneata, D.}, \bibinfo{author}{Verbeek, J.},
  \bibinfo{author}{Schmid, C.}, \bibinfo{year}{2013}.
\newblock \bibinfo{title}{Action and event recognition with fisher vectors on a
  compact feature set}, in: \bibinfo{booktitle}{Proceedings of the IEEE
  Conference on Computer Vision and Pattern Recognition}, pp.
  \bibinfo{pages}{1817--1824}.
\bibitem[{Phan et~al.(2015)Phan, Le and Satoh}]{phan2015multimedia}
\bibinfo{author}{Phan, S.}, \bibinfo{author}{Le, D.D.}, \bibinfo{author}{Satoh,
  S.}, \bibinfo{year}{2015}.
\newblock \bibinfo{title}{Multimedia event detection using event-driven
  multiple instance learning}, in: \bibinfo{booktitle}{Proceedings of the 23rd
  ACM International Conference on Multimedia}, pp. \bibinfo{pages}{1255--1258}.
\bibitem[{Rabiner and Schafer(2007)}]{rabiner2007introduction}
\bibinfo{author}{Rabiner, L.}, \bibinfo{author}{Schafer, R.},
  \bibinfo{year}{2007}.
\newblock \bibinfo{title}{Introduction to digital speech processing}.
\newblock \bibinfo{journal}{Foundations and Trends in Signal Processing}
  \bibinfo{volume}{1}, \bibinfo{pages}{1--194}.
\bibitem[{Ren et~al.(2021)Ren, Xiao, Chang, Huang, Li, Chen and
  Wang}]{RenXCHLCW21}
\bibinfo{author}{Ren, P.}, \bibinfo{author}{Xiao, Y.}, \bibinfo{author}{Chang,
  X.}, \bibinfo{author}{Huang, P.}, \bibinfo{author}{Li, Z.},
  \bibinfo{author}{Chen, X.}, \bibinfo{author}{Wang, X.}, \bibinfo{year}{2021}.
\newblock \bibinfo{title}{A comprehensive survey of neural architecture search:
  Challenges and solutions}.
\newblock \bibinfo{journal}{{ACM} Comput. Surv.} \bibinfo{volume}{54},
  \bibinfo{pages}{76:1--76:34}.
\bibitem[{S{\'a}nchez et~al.(2013)S{\'a}nchez, Perronnin, Mensink and
  Verbeek}]{sanchez2013image}
\bibinfo{author}{S{\'a}nchez, J.}, \bibinfo{author}{Perronnin, F.},
  \bibinfo{author}{Mensink, T.}, \bibinfo{author}{Verbeek, J.},
  \bibinfo{year}{2013}.
\newblock \bibinfo{title}{Image classification with the fisher vector: Theory
  and practice}.
\newblock \bibinfo{journal}{International Journal of Computer Vision}
  \bibinfo{volume}{105}, \bibinfo{pages}{222--245}.
\bibitem[{SanMiguel and Mart{\'\i}nez(2012)}]{sanmiguel2012semantic}
\bibinfo{author}{SanMiguel, J.C.}, \bibinfo{author}{Mart{\'\i}nez, J.M.},
  \bibinfo{year}{2012}.
\newblock \bibinfo{title}{A semantic-based probabilistic approach for real-time
  video event recognition}.
\newblock \bibinfo{journal}{Computer Vision and Image Understanding}
  \bibinfo{volume}{116}, \bibinfo{pages}{937--952}.
\bibitem[{{Shen} et~al.(2008){Shen}, {Tao} and {Li}}]{shen2008modality}
\bibinfo{author}{{Shen}, J.}, \bibinfo{author}{{Tao}, D.},
  \bibinfo{author}{{Li}, X.}, \bibinfo{year}{2008}.
\newblock \bibinfo{title}{Modality mixture projections for semantic video event
  detection}.
\newblock \bibinfo{journal}{IEEE Transactions on Circuits and Systems for Video
  Technology} \bibinfo{volume}{18}, \bibinfo{pages}{1587--1596}.
\bibitem[{Simonyan and Zisserman(2015)}]{VGG16}
\bibinfo{author}{Simonyan, K.}, \bibinfo{author}{Zisserman, A.},
  \bibinfo{year}{2015}.
\newblock \bibinfo{title}{Very deep convolutional networks for large-scale
  image recognition}, in: \bibinfo{booktitle}{Proceedings of the 3th
  International Conference on Learning Representations}.
\bibitem[{Snoek and Smeulders(2010)}]{snoek2010visual}
\bibinfo{author}{Snoek, C.G.}, \bibinfo{author}{Smeulders, A.W.},
  \bibinfo{year}{2010}.
\newblock \bibinfo{title}{Visual-concept search solved?}
\newblock \bibinfo{journal}{Computer} \bibinfo{volume}{43},
  \bibinfo{pages}{76--78}.
\bibitem[{Song et~al.(2017)Song, Wu, Yu and Jia}]{song2017extracting}
\bibinfo{author}{Song, H.}, \bibinfo{author}{Wu, X.}, \bibinfo{author}{Yu, W.},
  \bibinfo{author}{Jia, Y.}, \bibinfo{year}{2017}.
\newblock \bibinfo{title}{Extracting key segments of videos for event detection
  by learning from web sources}.
\newblock \bibinfo{journal}{IEEE Transactions on Multimedia}
  \bibinfo{volume}{20}, \bibinfo{pages}{1088--1100}.
\bibitem[{Stein and McKenna(2017)}]{stein2017recognising}
\bibinfo{author}{Stein, S.}, \bibinfo{author}{McKenna, S.J.},
  \bibinfo{year}{2017}.
\newblock \bibinfo{title}{Recognising complex activities with histograms of
  relative tracklets}.
\newblock \bibinfo{journal}{Computer Vision and Image Understanding}
  \bibinfo{volume}{154}, \bibinfo{pages}{82--93}.
\bibitem[{Sun and Nevatia(2013)}]{sun2013large}
\bibinfo{author}{Sun, C.}, \bibinfo{author}{Nevatia, R.}, \bibinfo{year}{2013}.
\newblock \bibinfo{title}{Large-scale web video event classification by use of
  fisher vectors}, in: \bibinfo{booktitle}{Proceedings of IEEE Workshop on
  Applications of Computer Vision}, pp. \bibinfo{pages}{15--22}.
\bibitem[{Supancic and Ramanan(2013)}]{supancic2013self-paced}
\bibinfo{author}{Supancic, J.S.}, \bibinfo{author}{Ramanan, D.},
  \bibinfo{year}{2013}.
\newblock \bibinfo{title}{Self-paced learning for long-term tracking}, in:
  \bibinfo{booktitle}{Proceedings of the IEEE Conference on Computer Vision and
  Pattern Recognition}, pp. \bibinfo{pages}{2379--2386}.
\bibitem[{Tang et~al.(2012a)Tang, Fei-Fei and Koller}]{tang2012learning}
\bibinfo{author}{Tang, K.}, \bibinfo{author}{Fei-Fei, L.},
  \bibinfo{author}{Koller, D.}, \bibinfo{year}{2012}a.
\newblock \bibinfo{title}{Learning latent temporal structure for complex event
  detection}, in: \bibinfo{booktitle}{Proceedings of the IEEE Conference on
  Computer Vision and Pattern Recognition}, pp. \bibinfo{pages}{1250--1257}.
\bibitem[{Tang et~al.(2012b)Tang, Yang and Gao}]{tang2012self-paced}
\bibinfo{author}{Tang, Y.}, \bibinfo{author}{Yang, Y.B.}, \bibinfo{author}{Gao,
  Y.}, \bibinfo{year}{2012}b.
\newblock \bibinfo{title}{Self-paced dictionary learning for image
  classification}, in: \bibinfo{booktitle}{Proceedings of the 20th ACM
  International Conference on Multimedia}, pp. \bibinfo{pages}{833--836}.
\bibitem[{Tibo et~al.(2020)Tibo, Jaeger and Frasconi}]{tibo2020learning}
\bibinfo{author}{Tibo, A.}, \bibinfo{author}{Jaeger, M.},
  \bibinfo{author}{Frasconi, P.}, \bibinfo{year}{2020}.
\newblock \bibinfo{title}{Learning and interpreting multi-multi-instance
  learning networks}.
\newblock \bibinfo{journal}{Journal of Machine Learning Research}
  \bibinfo{volume}{21}, \bibinfo{pages}{1--60}.
\bibitem[{Tran et~al.(2015)Tran, Bourdev, Fergus, Torresani and
  Paluri}]{tran2015learning}
\bibinfo{author}{Tran, D.}, \bibinfo{author}{Bourdev, L.},
  \bibinfo{author}{Fergus, R.}, \bibinfo{author}{Torresani, L.},
  \bibinfo{author}{Paluri, M.}, \bibinfo{year}{2015}.
\newblock \bibinfo{title}{Learning spatiotemporal features with 3d
  convolutional networks}, in: \bibinfo{booktitle}{Proceedings of the IEEE
  International Conference on Computer Vision}, pp.
  \bibinfo{pages}{4489--4497}.
\bibitem[{Vahdat et~al.(2013)Vahdat, Cannons, Mori, Oh and
  Kim}]{vahdat2013compositional}
\bibinfo{author}{Vahdat, A.}, \bibinfo{author}{Cannons, K.},
  \bibinfo{author}{Mori, G.}, \bibinfo{author}{Oh, S.}, \bibinfo{author}{Kim,
  I.}, \bibinfo{year}{2013}.
\newblock \bibinfo{title}{Compositional models for video event detection: A
  multiple kernel learning latent variable approach}, in:
  \bibinfo{booktitle}{Proceedings of the IEEE International Conference on
  Computer Vision}, pp. \bibinfo{pages}{1185--1192}.
\bibitem[{Venugopalan et~al.(2015)Venugopalan, Rohrbach, Donahue, Mooney,
  Darrell and Saenko}]{Venugopalan2015Sequence}
\bibinfo{author}{Venugopalan, S.}, \bibinfo{author}{Rohrbach, M.},
  \bibinfo{author}{Donahue, J.}, \bibinfo{author}{Mooney, R.},
  \bibinfo{author}{Darrell, T.}, \bibinfo{author}{Saenko, K.},
  \bibinfo{year}{2015}.
\newblock \bibinfo{title}{Sequence to sequence -- video to text}, in:
  \bibinfo{booktitle}{Proceedings of the IEEE International Conference on
  Computer Vision}, pp. \bibinfo{pages}{4534--4542}.
\bibitem[{Wang et~al.(2013)Wang, Kl{\"a}ser, Schmid and Liu}]{wang2013dense}
\bibinfo{author}{Wang, H.}, \bibinfo{author}{Kl{\"a}ser, A.},
  \bibinfo{author}{Schmid, C.}, \bibinfo{author}{Liu, C.L.},
  \bibinfo{year}{2013}.
\newblock \bibinfo{title}{Dense trajectories and motion boundary descriptors
  for action recognition}.
\newblock \bibinfo{journal}{International Journal of Computer Vision}
  \bibinfo{volume}{103}, \bibinfo{pages}{60--79}.
\bibitem[{Wang and Schmid(2014)}]{Wang2014Action}
\bibinfo{author}{Wang, H.}, \bibinfo{author}{Schmid, C.}, \bibinfo{year}{2014}.
\newblock \bibinfo{title}{Action recognition with improved trajectories}, in:
  \bibinfo{booktitle}{Proceedings of the IEEE International Conference on
  Computer Vision}, pp. \bibinfo{pages}{3551--3558}.
\bibitem[{Wang et~al.(2019)Wang, Yan, Tang, Liu and Guo}]{wang2019bag}
\bibinfo{author}{Wang, X.}, \bibinfo{author}{Yan, Y.}, \bibinfo{author}{Tang,
  P.}, \bibinfo{author}{Liu, W.}, \bibinfo{author}{Guo, X.},
  \bibinfo{year}{2019}.
\newblock \bibinfo{title}{Bag similarity network for deep multi-instance
  learning}.
\newblock \bibinfo{journal}{Information Sciences} \bibinfo{volume}{504},
  \bibinfo{pages}{578--588}.
\bibitem[{Wu et~al.(2000)Wu, Kankanhalli, Lim and Hong}]{wu2006perspectives}
\bibinfo{author}{Wu, J.K.}, \bibinfo{author}{Kankanhalli, M.S.},
  \bibinfo{author}{Lim, J.H.}, \bibinfo{author}{Hong, D.},
  \bibinfo{year}{2000}.
\newblock \bibinfo{title}{Perspectives on Content-Based Multimedia Systems}.
\newblock \bibinfo{publisher}{Kluwer Academic, Hingham, MA}.
\bibitem[{Xian et~al.(2016)Xian, Rong, Yang and Tian}]{xian2016evaluation}
\bibinfo{author}{Xian, Y.}, \bibinfo{author}{Rong, X.}, \bibinfo{author}{Yang,
  X.}, \bibinfo{author}{Tian, Y.}, \bibinfo{year}{2016}.
\newblock \bibinfo{title}{Evaluation of low-level features for real-world
  surveillance event detection}.
\newblock \bibinfo{journal}{IEEE Transactions on Circuits and Systems for Video
  Technology} \bibinfo{volume}{27}, \bibinfo{pages}{624--634}.
\bibitem[{Xu et~al.(2015)Xu, Yang and Hauptmann}]{xu2015discriminative}
\bibinfo{author}{Xu, Z.}, \bibinfo{author}{Yang, Y.},
  \bibinfo{author}{Hauptmann, A.G.}, \bibinfo{year}{2015}.
\newblock \bibinfo{title}{A discriminative cnn video representation for event
  detection}, in: \bibinfo{booktitle}{Proceedings of the IEEE Conference on
  Computer Vision and Pattern Recognition}, pp. \bibinfo{pages}{1798--1807}.
\bibitem[{Yan et~al.(2021)Yan, Chang, Luo, Zheng, Zhang, Li and
  Nie}]{YanCLZZLN21}
\bibinfo{author}{Yan, C.}, \bibinfo{author}{Chang, X.}, \bibinfo{author}{Luo,
  M.}, \bibinfo{author}{Zheng, Q.}, \bibinfo{author}{Zhang, X.},
  \bibinfo{author}{Li, Z.}, \bibinfo{author}{Nie, F.}, \bibinfo{year}{2021}.
\newblock \bibinfo{title}{Self-weighted robust {LDA} for multiclass
  classification with edge classes}.
\newblock \bibinfo{journal}{{ACM} Trans. Intell. Syst. Technol.}
  \bibinfo{volume}{12}, \bibinfo{pages}{4:1--4:19}.
\bibitem[{Yan et~al.(2020)Yan, Zheng, Chang, Luo, Yeh and
  Hauptmann}]{YanZCLYH20}
\bibinfo{author}{Yan, C.}, \bibinfo{author}{Zheng, Q.}, \bibinfo{author}{Chang,
  X.}, \bibinfo{author}{Luo, M.}, \bibinfo{author}{Yeh, C.},
  \bibinfo{author}{Hauptmann, A.G.}, \bibinfo{year}{2020}.
\newblock \bibinfo{title}{Semantics-preserving graph propagation for zero-shot
  object detection}.
\newblock \bibinfo{journal}{{IEEE} Trans. Image Process.} \bibinfo{volume}{29},
  \bibinfo{pages}{8163--8176}.
\bibitem[{Yan et~al.(2015)Yan, Yang, Meng, Liu, Tong, Hauptmann and
  Sebe}]{yan2015event}
\bibinfo{author}{Yan, Y.}, \bibinfo{author}{Yang, Y.}, \bibinfo{author}{Meng,
  D.}, \bibinfo{author}{Liu, G.}, \bibinfo{author}{Tong, W.},
  \bibinfo{author}{Hauptmann, A.G.}, \bibinfo{author}{Sebe, N.},
  \bibinfo{year}{2015}.
\newblock \bibinfo{title}{Event oriented dictionary learning for complex event
  detection}.
\newblock \bibinfo{journal}{IEEE Transactions on Image Processing}
  \bibinfo{volume}{24}, \bibinfo{pages}{1867--1878}.
\bibitem[{Yuan et~al.(2021)Yuan, Chang, Huang, Liu and He}]{YuanCHLH21}
\bibinfo{author}{Yuan, D.}, \bibinfo{author}{Chang, X.},
  \bibinfo{author}{Huang, P.}, \bibinfo{author}{Liu, Q.}, \bibinfo{author}{He,
  Z.}, \bibinfo{year}{2021}.
\newblock \bibinfo{title}{Self-supervised deep correlation tracking}.
\newblock \bibinfo{journal}{{IEEE} Trans. Image Process.} \bibinfo{volume}{30},
  \bibinfo{pages}{976--985}.
\bibitem[{Zha et~al.(2015)Zha, Luisier, Andrews, Srivastava and
  Salakhutdinov}]{zha2015exploiting}
\bibinfo{author}{Zha, S.}, \bibinfo{author}{Luisier, F.},
  \bibinfo{author}{Andrews, W.}, \bibinfo{author}{Srivastava, N.},
  \bibinfo{author}{Salakhutdinov, R.}, \bibinfo{year}{2015}.
\newblock \bibinfo{title}{Exploiting image-trained cnn architectures for
  unconstrained video classification}, in: \bibinfo{booktitle}{Proceedings of
  the 26-th British Machine Vision Conference}, pp.
  \bibinfo{pages}{1097--1105}.
\bibitem[{Zhan et~al.(2019)Zhan, Chang, Guan, Chen, Ma and Yang}]{ZhanCGCMY19}
\bibinfo{author}{Zhan, K.}, \bibinfo{author}{Chang, X.}, \bibinfo{author}{Guan,
  J.}, \bibinfo{author}{Chen, L.}, \bibinfo{author}{Ma, Z.},
  \bibinfo{author}{Yang, Y.}, \bibinfo{year}{2019}.
\newblock \bibinfo{title}{Adaptive structure discovery for multimedia analysis
  using multiple features}.
\newblock \bibinfo{journal}{{IEEE} Trans. Cybern.} \bibinfo{volume}{49},
  \bibinfo{pages}{1826--1834}.
\bibitem[{Zhang et~al.(2006)Zhang, Platt and Viola}]{zhang2006multiple}
\bibinfo{author}{Zhang, C.}, \bibinfo{author}{Platt, J.C.},
  \bibinfo{author}{Viola, P.A.}, \bibinfo{year}{2006}.
\newblock \bibinfo{title}{Multiple instance boosting for object detection}, in:
  \bibinfo{booktitle}{Proceedings of the Advances in Neural Information
  Processing Systems}, pp. \bibinfo{pages}{1417--1424}.
\bibitem[{Zhang et~al.(2017a)Zhang, Han, Jiang, Ye and Chang}]{ZhangHJYC17}
\bibinfo{author}{Zhang, D.}, \bibinfo{author}{Han, J.}, \bibinfo{author}{Jiang,
  L.}, \bibinfo{author}{Ye, S.}, \bibinfo{author}{Chang, X.},
  \bibinfo{year}{2017}a.
\newblock \bibinfo{title}{Revealing event saliency in unconstrained video
  collection}.
\newblock \bibinfo{journal}{{IEEE} Trans. Image Process.} \bibinfo{volume}{26},
  \bibinfo{pages}{1746--1758}.
\bibitem[{Zhang et~al.(2017b)Zhang, Meng and Han}]{zhang2017co-saliency}
\bibinfo{author}{Zhang, D.}, \bibinfo{author}{Meng, D.}, \bibinfo{author}{Han,
  J.}, \bibinfo{year}{2017}b.
\newblock \bibinfo{title}{Co-saliency detection via a self-paced
  multiple-instance learning framework}.
\newblock \bibinfo{journal}{IEEE Transactions on Pattern Analysis and Machine
  Intelligence} \bibinfo{volume}{39}, \bibinfo{pages}{865--878}.
\bibitem[{Zhang et~al.(2015)Zhang, Meng, Li, Jiang, Zhao and
  Han}]{zhang2015self}
\bibinfo{author}{Zhang, D.}, \bibinfo{author}{Meng, D.}, \bibinfo{author}{Li,
  C.}, \bibinfo{author}{Jiang, L.}, \bibinfo{author}{Zhao, Q.},
  \bibinfo{author}{Han, J.}, \bibinfo{year}{2015}.
\newblock \bibinfo{title}{A self-paced multiple-instance learning framework for
  co-saliency detection}, in: \bibinfo{booktitle}{Proceedings of the IEEE
  International Conference on Computer Vision}, pp. \bibinfo{pages}{594--602}.
\bibitem[{Zhang et~al.(2020a)Zhang, Chang, Liu, Luo, Prakash and
  Hauptmann}]{ZhangCLLPH20}
\bibinfo{author}{Zhang, L.}, \bibinfo{author}{Chang, X.}, \bibinfo{author}{Liu,
  J.}, \bibinfo{author}{Luo, M.}, \bibinfo{author}{Prakash, M.},
  \bibinfo{author}{Hauptmann, A.G.}, \bibinfo{year}{2020}a.
\newblock \bibinfo{title}{Few-shot activity recognition with cross-modal memory
  network}.
\newblock \bibinfo{journal}{Pattern Recognit.} \bibinfo{volume}{108},
  \bibinfo{pages}{107348}.
\bibitem[{Zhang et~al.(2018)Zhang, Liu, Luo, Chang and Zheng}]{Zhang0LCZ18}
\bibinfo{author}{Zhang, L.}, \bibinfo{author}{Liu, J.}, \bibinfo{author}{Luo,
  M.}, \bibinfo{author}{Chang, X.}, \bibinfo{author}{Zheng, Q.},
  \bibinfo{year}{2018}.
\newblock \bibinfo{title}{Deep semisupervised zero-shot learning with maximum
  mean discrepancy}.
\newblock \bibinfo{journal}{Neural Comput.} \bibinfo{volume}{30}.
\bibitem[{Zhang et~al.(2020b)Zhang, Luo, Liu, Chang, Yang and
  Hauptmann}]{ZhangLLCYH20}
\bibinfo{author}{Zhang, L.}, \bibinfo{author}{Luo, M.}, \bibinfo{author}{Liu,
  J.}, \bibinfo{author}{Chang, X.}, \bibinfo{author}{Yang, Y.},
  \bibinfo{author}{Hauptmann, A.G.}, \bibinfo{year}{2020}b.
\newblock \bibinfo{title}{Deep top-{\textdollar}k{\textdollar} ranking for
  image-sentence matching}.
\newblock \bibinfo{journal}{{IEEE} Trans. Multim.} \bibinfo{volume}{22},
  \bibinfo{pages}{775--785}.
\bibitem[{Zhang et~al.(2002)Zhang, Goldman, Yu and Fritts}]{zhang2002content}
\bibinfo{author}{Zhang, Q.}, \bibinfo{author}{Goldman, S.A.},
  \bibinfo{author}{Yu, W.}, \bibinfo{author}{Fritts, J.E.},
  \bibinfo{year}{2002}.
\newblock \bibinfo{title}{Content-based image retrieval using multiple-instance
  learning}, in: \bibinfo{booktitle}{Proceedings of the Nineteenth
  International Conference on Machine Learning}, pp. \bibinfo{pages}{682--689}.
\bibitem[{Zhao et~al.(2015)Zhao, Meng, Jiang, Xie, Xu and
  Hauptmann}]{zhao2015self-paced}
\bibinfo{author}{Zhao, Q.}, \bibinfo{author}{Meng, D.}, \bibinfo{author}{Jiang,
  L.}, \bibinfo{author}{Xie, Q.}, \bibinfo{author}{Xu, Z.},
  \bibinfo{author}{Hauptmann, A.G.}, \bibinfo{year}{2015}.
\newblock \bibinfo{title}{Self-paced learning for matrix factorization}, in:
  \bibinfo{booktitle}{Proceedings of the 29th AAAI Conference on Artificial
  Intelligence}, pp. \bibinfo{pages}{3196--3202}.
\bibitem[{Zhao et~al.(2018)Zhao, Xiang and Su}]{zhao2018complex}
\bibinfo{author}{Zhao, Z.}, \bibinfo{author}{Xiang, R.}, \bibinfo{author}{Su,
  F.}, \bibinfo{year}{2018}.
\newblock \bibinfo{title}{Complex event detection via attention-based video
  representation and classification}.
\newblock \bibinfo{journal}{Multimedia Tools and Applications}
  \bibinfo{volume}{77}, \bibinfo{pages}{3209--3227}.
\bibitem[{Zhou et~al.(2018)Zhou, Wang, Meng, Xin, Li, Gong and
  Zheng}]{zhou2018deep}
\bibinfo{author}{Zhou, S.}, \bibinfo{author}{Wang, J.}, \bibinfo{author}{Meng,
  D.}, \bibinfo{author}{Xin, X.}, \bibinfo{author}{Li, Y.},
  \bibinfo{author}{Gong, Y.}, \bibinfo{author}{Zheng, N.},
  \bibinfo{year}{2018}.
\newblock \bibinfo{title}{Deep self-paced learning for person
  re-identification}.
\newblock \bibinfo{journal}{Pattern Recognition} \bibinfo{volume}{76},
  \bibinfo{pages}{739--751}.
\bibitem[{Zhou et~al.(2009)Zhou, Sun and Li}]{zhou2009multi}
\bibinfo{author}{Zhou, Z.H.}, \bibinfo{author}{Sun, Y.Y.}, \bibinfo{author}{Li,
  Y.F.}, \bibinfo{year}{2009}.
\newblock \bibinfo{title}{Multi-instance learning by treating instances as
  non-iid samples}, in: \bibinfo{booktitle}{Proceedings of the 26th
  International Conference on Machine Learning}, pp.
  \bibinfo{pages}{1249--1256}.

\end{thebibliography}



\end{document}